\documentclass[sigconf, 9pt]{acmart}

\usepackage[english]{babel}
\usepackage{blindtext}
\usepackage{caption}
\usepackage{subcaption}
\usepackage{color}
\usepackage{tabularx}
\usepackage{multirow}
\usepackage{multicol}
\usepackage{graphicx}
\usepackage{tabularray}
\usepackage{algorithm}          
\usepackage{algpseudocode}  

\usepackage{float}
\usepackage{xcolor} 
\usepackage{todonotes}
\usepackage{array} 
\usepackage{makecell} 
\usepackage{booktabs} 
\usepackage{geometry}
\usepackage{microtype}
\usepackage{bbding}

\newcommand{\name}{MoViD\xspace}

\begin{document}

\acmYear{2026}\copyrightyear{2026}
\setcopyright{cc}
\setcctype[4.0]{by}
\acmConference[SenSys '26]{ACM/IEEE International Conference on Embedded Artificial Intelligence and Sensing Systems}{May 11--14, 2026}{Saint Malo, France}
\acmBooktitle{ACM/IEEE International Conference on Embedded Artificial Intelligence and Sensing Systems (SenSys '26), May 11--14, 2026, Saint Malo, France}
\acmDOI{10.1145/3774906.3802786}
\acmISBN{979-8-4007-2309-4/26/05}

\title{MoViD: View-Invariant 3D Human Pose Estimation via Motion-View Disentanglement}

\author{Yejia Liu\textsuperscript{*}, Hengle Jiang\textsuperscript{*}, Haoxian Liu, Runxi Huang, Xiaomin Ouyang\textsuperscript{\dag}}
\affiliation{
  \institution{Department of Computer Science and Engineering, Hong Kong University of Science and Technology}
  \country{}
}
\email{{yliutb, hjiangbg, hliueu, rhuangbj}@connect.ust.hk, xmouyang@cse.ust.hk}
\thanks{\textsuperscript{*}These authors contributed equally. \textsuperscript{\dag}Corresponding author.}









\begin{abstract}
3D human pose estimation is a key enabling technology for applications such as healthcare monitoring, human-robot collaboration, and immersive gaming, but real-world deployment remains challenged by viewpoint variations. Existing methods struggle to generalize to unseen camera viewpoints, require large amounts of training data, and suffer from high inference latency. We propose MoViD, a viewpoint-invariant 3D human pose estimation framework that disentangles viewpoint information from motion features. The key idea is to extract viewpoint information from intermediate pose features and leverage it to enhance both the robustness and efficiency of pose estimation. MoViD introduces a view estimator that models key joint relationships to predict viewpoint information, and an orthogonal projection module to disentangle motion and view features, further enhanced through physics-grounded contrastive alignment across views. For real-time edge deployment, MoViD employs a frame-by-frame inference pipeline with a view-aware strategy that adaptively activates flip refinement based on the estimated viewpoint. Evaluations on nine public datasets and newly collected multiview UAV and gait analysis datasets show that MoViD reduces pose estimation error by over 24.2\% compared to state-of-the-art methods, maintains robust performance under severe occlusions with 60\% less training data, and achieves real-time inference at 15 FPS on NVIDIA edge devices.\footnote{Code available at: \url{https://github.com/HKUST-MINSys-Lab/MoViD}. Video demonstration available at: \url{https://youtu.be/L4Bx_LvPXB8}.}
\end{abstract}

\maketitle

 \begin{figure}
    \centering
     \setlength{\abovecaptionskip}{0.cm}
    \setlength{\belowcaptionskip}{0.cm}
    \includegraphics[width=1\linewidth]{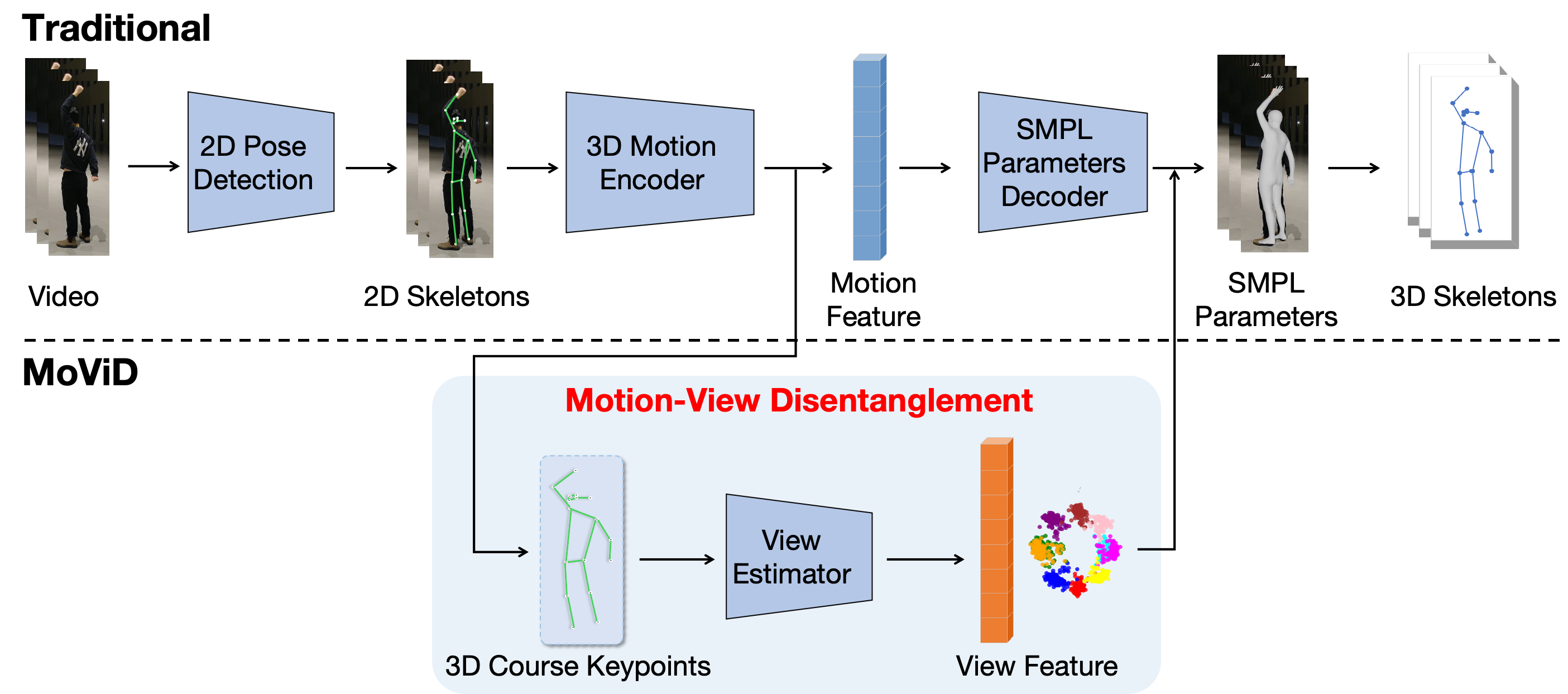}
    \caption{Comparison of the traditional framework for 3D human pose estimation and our proposed approach \name. The key idea of \name is to extract and leverage viewpoint information for view-invariant pose estimation. 
    }
    \label{fig:pipeline}
\end{figure}

\section{Introduction}

 
 
 
 
 
 

Human pose estimation from visual data is very important for understanding human motion for a wide range of applications, such as gait analysis \cite{stenum2021applications}, AI-driven fitness coaching \cite{hannan2021portable}, human-robot collaboration \cite{LIU2024human-robot} and daily behavior monitoring for chronic disease management \cite{ouyang2024admarker}.
In real-world human pose estimation applications, cameras are often deployed in homes or mounted on robots to capture video and estimate human poses in real time. For example, in smart home environments, vision-based monitoring systems can track the posture and movements of elderly residents to detect falls or mobility impairments, enabling timely alerts and assistance \cite{falldetection}. Figure~\ref{fig:pipeline} illustrates the standard procedure for 3D human pose estimation from video data, which typically involves two stages. First, a 2D pose estimation model, often based on convolutional neural networks (CNNs) or transformers, detects human keypoints (e.g., elbows, knees, hips) in individual video frames. Next, a 3D pose lifting module, such as a graph convolutional network (GCN) or temporal transformer, maps these 2D keypoints into the SMPL parameter space \cite{SMPL} to reconstruct the full 3D human mesh. Finally, a set of predefined joint locations is extracted from the SMPL mesh vertices to obtain the corresponding 3D keypoints used for downstream analysis.


However, the reliability of human pose estimation systems is often compromised by viewpoint variations in real-world deployments. For instance, cameras in residential environments may be installed at different angles and heights due to differences in home layout, furniture arrangement, or mounting constraints. Similarly, vision systems mounted on mobile platforms such as robots or unmanned aerial vehicles (UAVs) experience continuous changes in viewpoint as they move through the environment. The shifting camera perspectives can significantly degrade pose estimation accuracy due to several key challenges.
First, changes in viewpoint introduce complex \emph{visual distortions} like foreshortening, where limbs appear shorter when pointing toward the camera, or even occlusions. This disrupts spatial consistency and yields inaccurate depth estimates, degrading 3D reconstruction. Second, diverse actions further complicate pose estimation, as natural movements (e.g., arm swings) may be misinterpreted as viewpoint shifts, making actual motion harder to distinguish from camera-induced variation.
Moreover, many pose estimation pipelines incur substantial overhead because they process video sequences, limiting real-time deployment.

Unfortunately, existing methods for 3D pose estimation struggle to handle viewpoint variations effectively. Some approaches attempt to improve diversity of training data by incorporating data from multiple views \cite{WEPDTOF-Pose, NToP}, or synthetic data generated from parametric models and rendered scenes \cite{SURREAL}. However, such data-driven approaches rely heavily on the training data distribution, often failing to generalize to unseen viewpoints. Other methods aim to achieve viewpoint robustness by extracting body-centric features such as dense correspondence maps \cite{Song2020Pose2Pose, Guler2018DensePose}, or leveraging 3D priors such as the SMPL model \cite{SMPL}. However, these techniques tend to degrade under extreme viewing angles \cite{Shahroudy2016NTU}, and often incur high computational costs. Recently, some approaches are focused on learning viewpoint-invariant features through contrastive learning \cite{Men2023FocalizedCV} or adversarial learning \cite{dong2022viewfool, Ruan2023TowardsVV}, but typically require extensive training with 3D supervision \cite{mildenhall2020nerf}, and face scalability issues in complex viewpoint scenarios \cite{Ruan2024OmniviewTuningBV}.


In this paper, we propose \name, a new framework for viewpoint-invariant human pose estimation through explicitly estimating and disentangling viewpoint information from motion features. Our key insight is that intermediate features used for 3D keypoint prediction implicitly encode viewpoint information. While these features may not be optimal for final pose estimation, they contain rich geometric information sufficient for accurate view inference. To exploit this, we design a view estimator that predicts viewpoint information directly from coarse 3D keypoint features through explicit modeling spatial relationships across joints. The estimated viewpoint information is then used to guide view-invariant skeleton extraction. Specifically, we design an orthogonal feature projection module that disentangles motion and view features by enforcing their orthogonality, effectively normalizing perspective distortions and ensuring motion features remain consistent across varying viewpoints. Such explicit geometric disentanglement is fundamentally different from and more efficient than prior implicit approaches based on data augmentation~\cite{mhformer} or adversarial training~\cite{repnet}. To further enhance robustness of cross-view features, we incorporate an integrated physics-enhanced contrastive learning mechanism that aligns motion features across views using contrastive loss and leverages SMPL-derived motion dynamics to impose physical constraints that further reinforce feature consistency. Finally, to enable real-time inference on edge devices, we design a frame-by-frame inference pipeline with adaptive activation of a flip refinement module. Instead of processing entire video sequences and applying the computation-intensive refinement uniformly, \name processes each frame individually and selectively activates the refinement module based on viewpoint estimates. Such design allows the model to adapt precisely to viewpoint changes, minimizing unnecessary computation to reduce inference latency while maintaining high accuracy.


We extensively evaluate \name using nine public datasets that capture various human activities and complex human-scene interactions across up to 10 camera viewpoints, collected in both indoor and outdoor environments. In addition, we collected two new datasets: a multiview UAV dataset captured under fast and continuous viewpoint changes, and a multimodal dataset for fine-grained gait analysis that includes synchronized data from RGB, infrared (IR), depth, 3D skeleton, and text prompts. The results show that \name reduces pose estimation error by over 28\%, while maintaining robust performance under severe occlusions with 60\% less training data. It also achieves real-time inference at 15 FPS on edge devices like Nvidia Jetson Orin NX, demonstrating its efficiency for practical deployment.

Our key contributions include:
\begin{itemize}

\item We conduct an in-depth analysis of the impact of viewpoint variations on 3D pose estimation and explore the potential of leveraging viewpoint information to improve performance.
\item Based on the key observations, we propose \name, a new framework that extracts viewpoint information via a dedicated view estimator and employs an orthogonal feature projection module to explicitly disentangle motion and view features for robust pose estimation.
\item \name incorporates a physics-enhanced contrastive alignment mechanism to enforce view-invariant motion representations across viewpoints, and features a frame-by-frame processing pipeline with adaptive module activation to reduce inference latency.
\item Extensive evaluations on public and two new self-collected datasets show that \name consistently outperforms state-of-the-art methods in pose estimation across diverse viewpoints with less training data, achieve robust performance under different conditions and incurs lower inference latency.

\end{itemize}

\section{Related Work}

\textbf{Viewpoint-Invariant 3D Human Pose Estimation}.
3D human pose estimation from visual data has gained increasing attention for applications like gait analysis \cite{stenum2021applications}, human-robot collaboration \cite{LIU2024human-robot} and daily behavior monitoring for chronic disease management \cite{ouyang2024admarker}. 
However, robust pose estimation across viewpoints remains challenging. 
Some approaches increase the diversity of training data by incorporating data from multiple views \cite{WEPDTOF-Pose, NToP}, or synthetic data generated from parametric models \cite{SURREAL}. However, these data-driven approaches often fail to generalize to unseen viewpoints \cite{PanopTOP, ObjectNet}. Other methods extract body features like dense correspondence maps \cite{Song2020Pose2Pose, Guler2018DensePose} or SMPL priors\cite{SMPL}, but degrade under extreme angles and are computationally expensive. Recent efforts employ contrastive \cite{Men2023FocalizedCV} or adversarial learning \cite{dong2022viewfool, Ruan2023TowardsVV} to learn viewpoint-invariant features, yet they require extensive 3D supervision and fail to scale in complex scenarios. Although approaches based on coordinate transformation ~\cite{Nie2019SPM, Li2022SimCC} can efficiently localize joints, they depend on camera axis information and lack generalization to new angles.
In contrast, \name extracts and disentangles viewpoint information from motion features for robust, data-efficient pose estimation under extreme viewing angles. It can also reduce latency through adaptive inference at runtime. 
Unlike prior viewpoint-aware methods~\cite{Ghezelghieh2016CNN3DV,Wang2020PredictingCV} that treat viewpoint as supplementary information---either as an additional input feature or as an auxiliary training signal---\name uses the predicted viewpoint to explicitly disentangle motion features from view-dependent features via Gram-Schmidt projection, ensuring the resulting motion representation is view-invariant by construction.

\textbf{Robust Vision Systems}.
Designing vision systems that can generalize across different environments is crucial for real-world deployment. In tasks like object recognition and tracking, model robustness is often improved through data augmentation \cite{ObjectNet} or synthetic rendering techniques such as NeRF \cite{mildenhall2020nerf}. However, these approaches usually suffer from domain gaps and poor scalability. 
For example, ObjectNet \cite{ObjectNet} shows that achieving comprehensive viewpoint coverage requires exponentially more data. 
Techniques like adversarial framework or adapting vision-language models can also improve robustness but are computationally intensive. For example, VIAT \cite{Ruan2023TowardsVV} requires over 400 GPU hours for 1,000 training samples. 
Moreover, extending these techniques to pose estimation is particularly challenging. Unlike object recognition, pose estimation requires precise reasoning over articulated joints, making it highly sensitive to viewpoint variations and occlusions. \name tackles this by explicitly modeling skeletal geometry to extract viewpoint-invariant features, leveraging kinematic structure rather than relying on data-intensive augmentation, enabling robust performance with less training data.

\section{A Motivation Study}
\label{sec:motivate_study}

In this section, we evaluate the performance of pose estimation across different viewpoints and explore the potential of leveraging viewpoint information. The key insights from these results motivate the design of \name.


\subsection{Impact of Viewpoint Variations}
\label{sec:impact_viewpoint}

In real-world scenarios, camera perspectives vary with deployment constraints or platform mobility. We investigate how viewpoint variations affect 3D human pose estimation accuracy. 


Specifically, we evaluate the performance of WHAM \cite{shin2023wham}, a state-of-the-art 3D human pose estimation model, using the HuMMan dataset \cite{cai2022humman}, which captures diverse human actions from 10 distinct camera viewpoints. 
Following standard practice, the pose estimation model is trained on five widely-used datasets: 3DPW \cite{3dpw}, Human3.6M \cite{Ionescu2014Human36MLS}, MPII-INF-3DHP \cite{mono-3dhp2017}, InstaVariety \cite{InstaVariety}, and AMASS \cite{AMASS:2019}.
While the training data includes various viewing angles, it is dominated by frontal views. In contrast, the evaluation dataset HuMMan features 8 horizontal viewpoints spaced at 45-degree intervals (0°, 45°, 90°, 135°, 180°, 225°, 270°, 315°) and 2 elevated side views. These side viewpoints are underrepresented in the training data, which may contribute to performance discrepancies.

\begin{figure}
    \centering
     \setlength{\abovecaptionskip}{0.cm}
    \setlength{\belowcaptionskip}{0.cm}
    \includegraphics[width=1\linewidth]{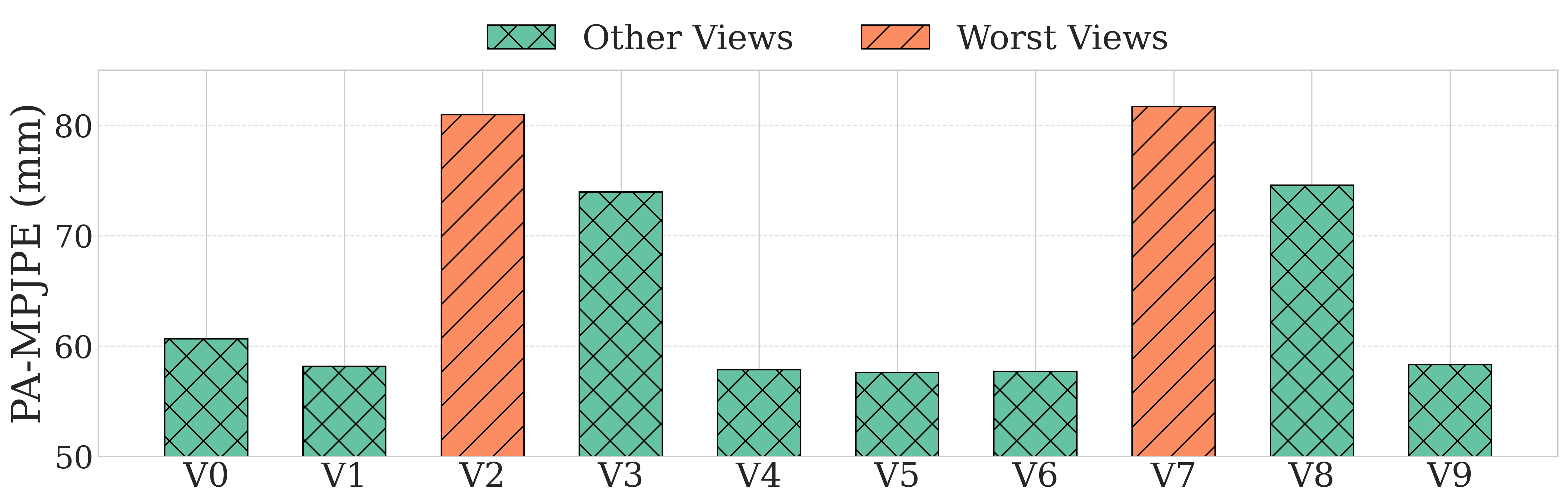}
    \caption{Pose estimation errors across different views.
    }
    \label{fig:cross_view}
\end{figure}

We evaluate pose estimation performance using the PA-MPJPE metric (lower is better) \cite{sun2018integral}, which measures mean per-joint position error using procrustes analysis. 
Figure~\ref{fig:cross_view} shows pose estimation errors across ten camera viewpoints (View 1–10), with the worst-performing views highlighted in orange. First, the results reveal substantial performance variation across views, ranging from 57.6 mm to 81.7 mm, which shows the impact of camera perspective on pose estimation accuracy. Moreover, several viewpoints, such as View 2 and View 7 (the elevated side viewpoints),  consistently yield higher errors than other views, indicating a strong correlation between viewpoint and performance degradation. These results show the performance of pose estimation varies significantly with camera viewpoints, especially under challenging perspectives. Therefore, developing viewpoint-robust methods is essential for ensuring reliable performance across diverse camera perspectives.



    

\subsection{Potential of Viewpoint Estimation}
To mitigate the impact of viewpoint variation, we explore incorporating viewpoint information during the process of pose estimation.

\subsubsection{Embedded viewpoint information in pose features. } 
\begin{figure}
     \setlength{\abovecaptionskip}{0.cm}
    \setlength{\belowcaptionskip}{-0cm}
    \centering
    \begin{minipage}{.43\linewidth}
    \setlength{\abovecaptionskip}{0.cm}
    \setlength{\belowcaptionskip}{-0.cm}
    \centering
	\includegraphics[width=\linewidth]{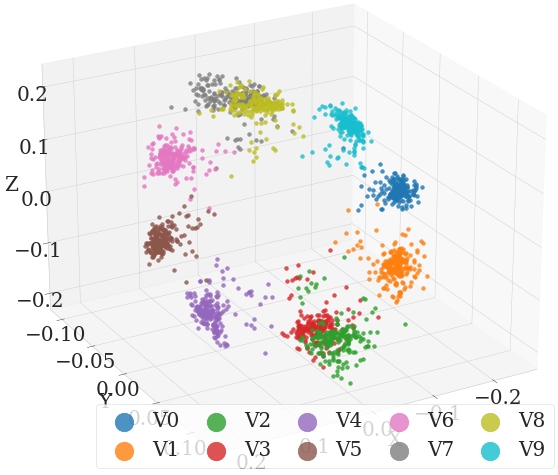}
	\caption{The intermediate pose features exhibit clear clustering patterns by viewpoint.
    }
	\label{fig:view_info}
    \end{minipage}
    \hspace{0.5pt}
  \begin{minipage}{.54\linewidth}
    \setlength{\abovecaptionskip}{0.cm}
    \setlength{\belowcaptionskip}{-0.cm}
    \centering
	\includegraphics[width=\linewidth]{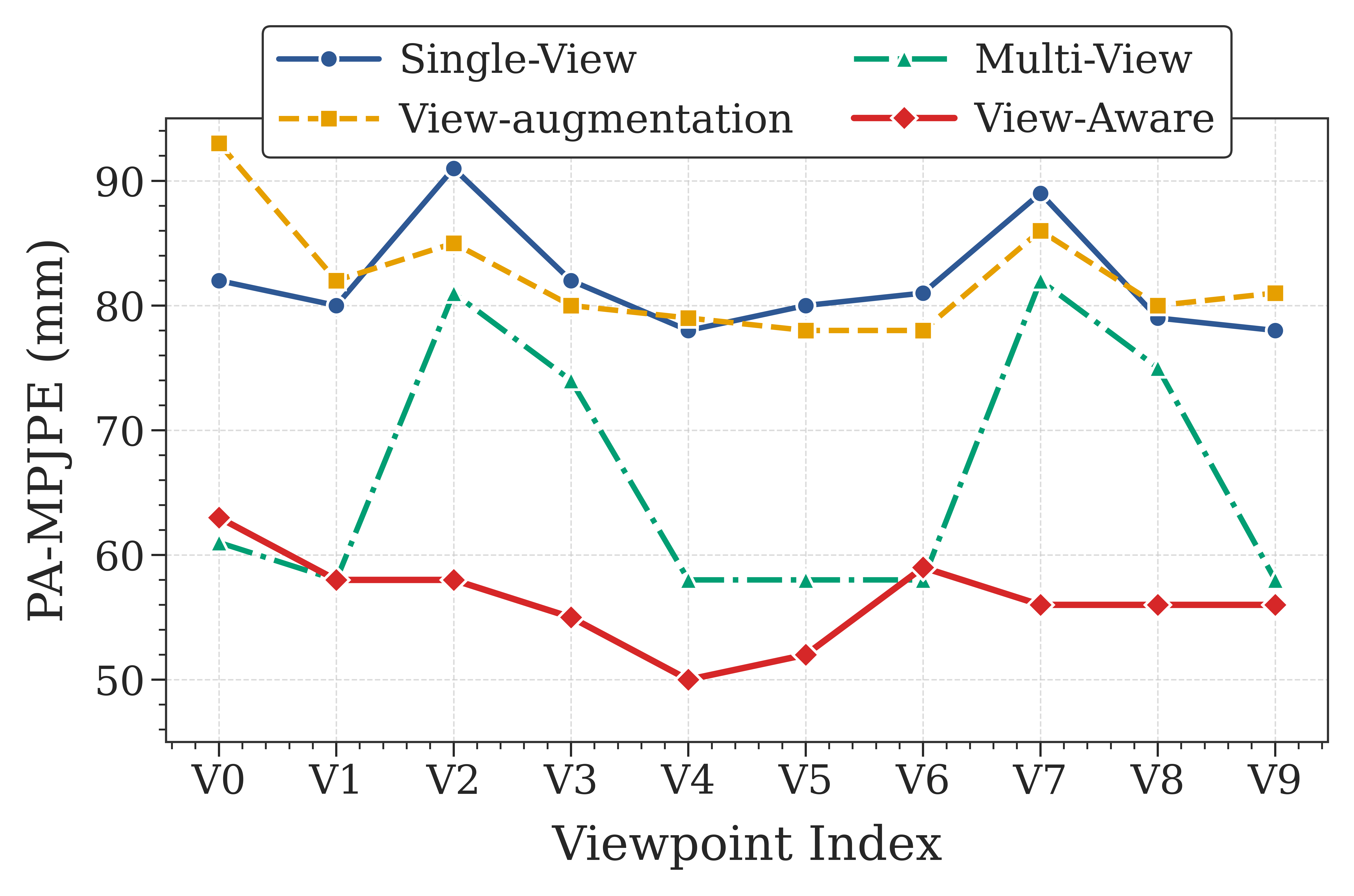}
	\caption{Incorporating view information (``View-aware'') reduces pose estimation errors across views.
 }
	\label{fig:baselines}
  \end{minipage}
\end{figure}

As illustrated in Figure~\ref{fig:pipeline}, variations in camera viewpoints can introduce discrepancies in 2D keypoint detection, resulting in distinct 3D spatial features. To investigate whether viewpoint information is encoded in intermediate pose features, we extract pose features from data captured across multiple camera views and project them into 3D using Principal Component Analysis (PCA). As shown in Figure~\ref{fig:view_info}, the features exhibit clear clustering patterns by viewpoint, indicating that viewpoint-related information is inherently embedded in the pose features. This insight can be exploited to improve robustness across diverse camera perspectives.

\subsubsection{Incorporating viewpoint information. } To motivate the use of viewpoint information in pose estimation, we design a method referred to as "View-aware", which integrates explicit viewpoint data into the pose estimation model via cross-attention with motion features. This approach enables the model to learn view-aware representations that improve robustness to viewpoint variation. We compare the ``View-aware'' approach with three representative baselines: ``Single-View'', ``Multi-View'', and ``View-Augmentation''. ``Single-View'' trains on a limited set of fixed views (View 1, 4, 6, and 9), while ``Multi-View'' uses a full 360° range of viewpoints. ``View-Augmentation'' extends Single-View by incorporating synthetic data from the AMASS dataset \cite{AMASS:2019} to improve generalization.



Figure~\ref{fig:baselines} shows PA-MPJPE errors (lower is better) across 10 camera viewpoints in the Human3.6M dataset \cite{Ionescu2014Human36MLS}. The ``Single-View'' approach performs well on its training views (View 1, 4, 6, 9) but generalizes poorly to unseen viewpoints. ``Multi-View'' achieves a moderate average error (66.3 mm) but suffers from high inconsistency, with a 14.9\% variation across views. Similarly, ``View-Augmentation'' yields an average error of 82.2 mm with large variation. This indicates that simply increasing view diversity in training data does not resolve viewpoint-induced feature heterogeneity. In contrast, the ``View-aware'' approach, which integrates explicit viewpoint information via cross-attention, achieves the lowest average error (56.3 mm) and strong consistency (6.1\% variation). Although this method requires viewpoint data that may be difficult to obtain in practice, it demonstrates the clear benefit of leveraging viewpoint information for robust pose estimation.




\subsection{Summary}
\label{sec:summary}

We now summarize the key findings from our motivation study.

\begin{itemize}
    \item Viewpoint variations significantly degrade pose estimation performance, especially under challenging views.
    \item Viewpoint-related information is \emph{inherently encoded in intermediate pose features}. Moreover, incorporating explicit viewpoint information into the model can significantly enhance performance across diverse camera perspectives.
\end{itemize}

However, explicit viewpoint data is often unavailable in real-world scenarios due to the lack of calibrated multi-camera setups or metadata specifying camera orientation. This motivates the need to estimate viewpoint information during the pose estimation process and effectively leverage it to enhance model performance. Another key advantage is that the inferred viewpoint information provides opportunities to enable adaptive inference for reducing computational overhead at runtime.


\section{System Overview}
\label{sec:overview}

We now introduce \name, a new framework for viewpoint-invariant human pose estimation through explicitly disentangling motion and view features. The key idea is to estimate viewpoint information from intermediate pose features and use it to improve pose estimation accuracy and efficiency. We first introduce the applications and challenges, then describe the system architecture.

\subsection{Applications and Challenges} 


\name is designed for a wide range of applications where vision systems operate in ambient environments or on mobile platforms to extract 3D human pose for various downstream tasks such as gait analysis \cite{stenum2021applications}, fitness coaching \cite{hannan2021portable}, human-robot collaboration \cite{LIU2024human-robot}, and daily behavior monitoring for chronic disease management \cite{ouyang2024admarker}. In these applications, camera viewpoints can vary significantly due to deployment constraints or platform mobility. \name can estimate viewpoint information during the pose estimation process to enhance robustness and computational efficiency in dynamic, real-world settings.

We now describe two typical application scenarios. First, in smart home or elderly care environments, cameras are often mounted on ceilings, walls, or corners, capturing human pose for tasks like activity recognition. However, variations in home layout and mounting constraints lead to inconsistent viewpoints between training and deployment. For instance, a fall detection system trained on frontal views may misclassify actions when observed from side or top-down perspectives \cite{yolofalldetection}. \name addresses this challenge by estimating viewpoint information during deployment, enabling more robust pose estimation across diverse real-world scenarios.
Second, mobile platforms such as robots or UAVs experience continuous viewpoint shifts as they navigate through dynamic environments. For example, in human-robot interaction, accurate pose interpretation across varying views is crucial, especially under extreme angles or occlusions during movement \cite{do2025skateformer}. Moreover, real-time human-robot interaction requires low-latency pose estimation to ensure timely  responses. In such scenarios, \name can estimate viewpoint information at runtime to enhance pose estimation accuracy and reduce prediction latency. In the following, we discuss these challenges in detail to motivate the design of \name.

\begin{figure}
    \centering
         \setlength{\abovecaptionskip}{0.cm}
    \setlength{\belowcaptionskip}{0.cm}
    \includegraphics[width=1\linewidth]{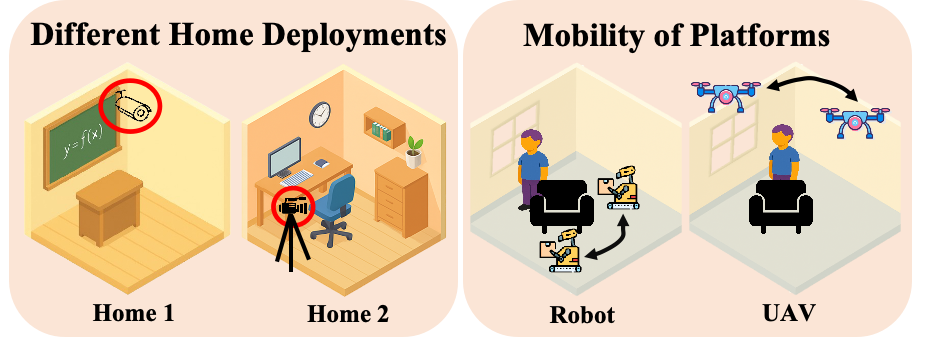}
    \caption{Typical application scenarios of \name: Activity recognition across diverse deployment settings and human-robot interaction during navigation. 
    }
    \label{fig:application}
\end{figure}

\begin{figure*}
    \centering
         \setlength{\abovecaptionskip}{0.cm}
    \setlength{\belowcaptionskip}{0.cm}
    \includegraphics[width=1\linewidth]{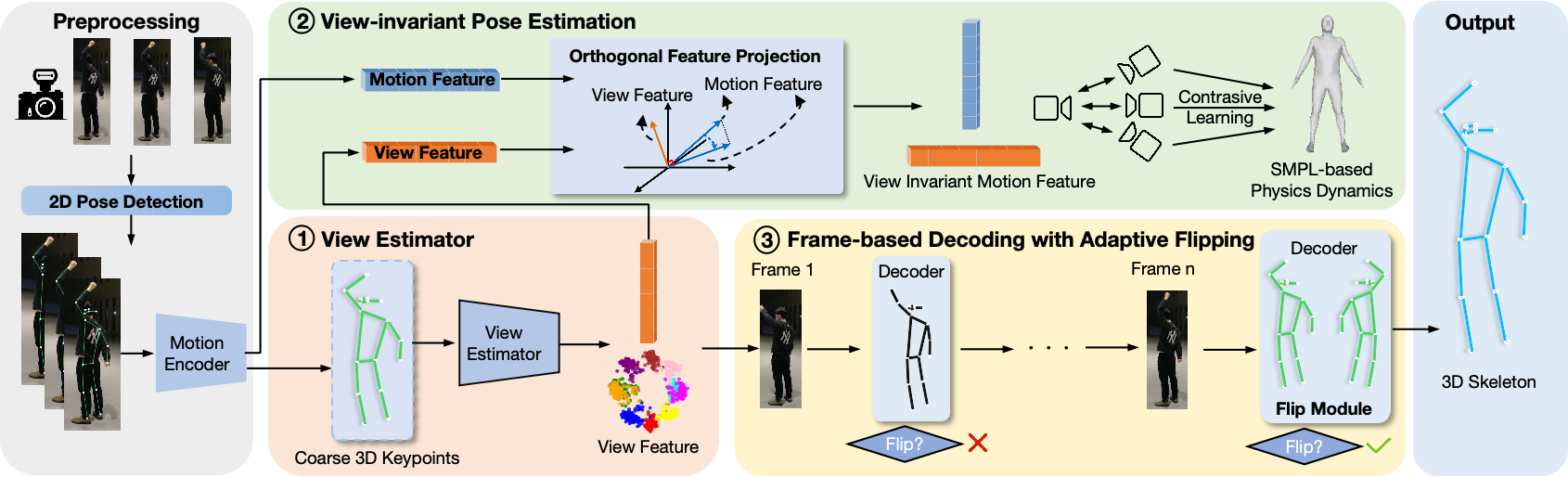}
    \caption{\name features a view estimator to predict viewpoint information from intermediate pose estimation features. This information is then leveraged to enable view-invariant pose estimation through view disentanglement and physics-enhanced contrastive learning, as well as enhancing frame-based adaptive inference to reduce latency. }
    \label{fig:overview}
\end{figure*}

\textbf{Generalization to unseen views and complex human actions.} A key challenge in 3D pose estimation is achieving robust generalization to unseen viewpoints and diverse human actions. Unseen viewpoint changes can introduce complex visual distortions, such as foreshortening and occlusions, which disrupt spatial consistency and often lead to inaccurate depth estimates. Moreover, natural movements in various human actions, such as arm swings, may be misinterpreted as viewpoint shifts, making it difficult for the model to distinguish between actual motion and camera-induced variation. These factors collectively degrade pose estimation accuracy in real-world scenarios.

\textbf{Limited training data.} In real-world applications, collecting and annotating large-scale 3D human pose data is often challenging, especially across diverse camera viewpoints and a wide range of human actions. Many environments, such as private homes or healthcare settings, impose constraints on data acquisition due to privacy concerns. Moreover, annotating ground-truth 3D human poses is significantly more labor-intensive than conventional tasks like activity recognition. Given these constraints, it is crucial to develop data-efficient methods that generalize well across various viewpoints without reliance on extensive labeled datasets.

\textbf{Significant inference latency.} 
Existing 3D pose estimation pipelines often suffer from  substantial inference overhead because they rely on \emph{processing video sequences} for accurate estimation. Moreover, standard approaches typically consist of multiple stages, including 2D keypoint detection, 2D-to-3D lifting, and temporal refinement, each requiring substantial processing time and resources. This makes real-time pose estimation challenging to achieve on resource-constrained devices, limiting deployment in time-sensitive applications such as human-robot interaction and video surveillance. Therefore, optimizing the pose estimation pipeline is essential to enable efficient, real-time execution on edge devices.

\subsection{System Architecture}

Motivated by Section~\ref{sec:motivate_study}, \name estimates viewpoint information from intermediate pose features to improve accuracy and reduce latency. Figure~\ref{sec:overview} shows the overall system architecture.



Specifically, \name first extracts coarse 3D keypoint features from 2D pose sequences using a lightweight motion encoder. While these features may not be optimal for final pose estimation, they contain sufficient geometric information for viewpoint prediction. To exploit this, we design a view estimator that predicts viewpoint information by explicitly modeling spatial relationships among joints. It uses a small joint set (e.g., hip and shoulder) across frames to compute spatial vectors, which are processed by a view encoder to generate viewpoint embeddings for viewpoint estimation.
The estimated viewpoint information is then used to guide view-invariant skeleton extraction. 
We design an orthogonal feature projection module to disentangle motion and view features through a Gram-Schmidt orthogonalization process that projects motion features into subspaces that are orthogonal to the view features. To further enhance the robustness of motion features, we incorporate a physics-enhanced contrastive alignment mechanism that aligns motion features across views using contrastive loss, and leverages SMPL-derived motion dynamics to provide physical constraints (e.g., acceleration and velocity) that further reinforce feature consistency. 
This preserves both view invariance and temporal consistency, even under noisy or incomplete visual observations.




To enable real-time edge inference, we use a frame-by-frame pipeline with adaptive activation of a flip refinement module. Rather than refining full video sequences uniformly, \name processes frames individually and triggers refinement only when viewpoint estimates indicate it is needed, reducing latency while maintaining accuracy.

\section{Design of \name}
\label{sec:Design}

\name is a new framework for view-robust 3D human motion analysis that disentangles viewpoint information from motion dynamics. 
It integrates three components: (1) a view estimator that predicts viewpoint information from intermediate skeleton features; (2) a view-invariant pose estimation module that explicitly disentangles motion and view via geometric orthogonal projection combined with physics-grounded supervision using SMPL-derived, view-independent anchors without discrete viewpoint labels; and (3) viewpoint-adaptive computation that selectively activates a flip refinement module based on estimated view difficulty for real-time edge deployment.

\subsection{View Estimator}
\label{sec:view_estimator}
As discussed in Section~\ref{sec:motivate_study}, viewpoint-related information is inherently embedded in pose estimation features. This is because variations in camera viewpoints introduce discrepancies in 2D keypoint detection, which in turn lead to distinct 3D spatial features \cite{PVNet}. Therefore, we design a view estimator to extract viewpoint information from intermediate 3D keypoint estimations, by explicitly modeling geometric relationships among joints and extracting view embeddings with a view encoder.

\subsubsection{Geometric feature extraction.} 
Given an input sequence $X$, the pose estimation model begins with a lightweight motion encoder that processes the input sequence to extract initial motion features $\mathbf{M}_{\text{init}}$  and coarse 3D keypoint predictions $\mathbf{K}_{3D}$. 
\begin{equation}
\mathbf{M}_{\text{init}}, \mathbf{K}_{3D} = Enc_{\text{motion}}(X).
\end{equation}
These 3D keypoint estimations capture the temporal dynamics and spatial configurations of human movement, but remain entangled with viewpoint information. 

Therefore, we first aim to extract geometric features with viewpoint information from the coarse 3D keypoints. Given a sequence of 3D keypoints $\mathbf{K}_{3D} \in \mathbb{R}^{T \times J \times 3}$, where $T$ denotes the number of frames, $J$ denotes the number of joints, and each joint is represented by $(x, y, z)$ coordinates, we extract geometric features from only a small set of key joints that best capture human body orientation while being minimally affected by motions~\cite{cite_skeleton_orientation}: the left/right hips ($\mathbf{k}_{\text{hip}_L}$, $\mathbf{k}_{\text{hip}_R}$) and the left/right shoulders ($\mathbf{k}_{\text{shoulder}_L}$, $\mathbf{k}_{\text{shoulder}_LR}$). 
The extremity and head joints from the coarse $\mathbf{K}_{3D}$ are less stable and more susceptible to occlusion, whereas hips and shoulders provide reliable geometric signals. Specifically, the left--right laterality of these joints inherently encodes the body's facing direction relative to the camera, while the lateral span between symmetric joints varies systematically with foreshortening under different viewpoints; the depth values (z-coordinates) capture perspective-dependent configurations, and the spatial vectors formed between hips and shoulders characterize the torso orientation, jointly offering a compact yet expressive representation of global body orientation and viewpoint.

\begin{figure}
    \centering
         \setlength{\abovecaptionskip}{0.cm}
    \setlength{\belowcaptionskip}{0.cm}
    \includegraphics[width=1\linewidth]{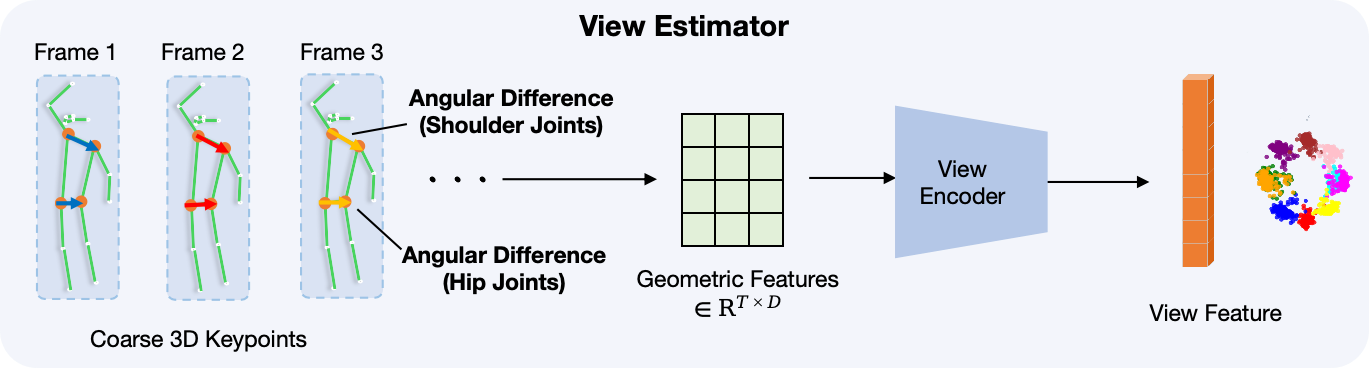}
\caption{Illustration of the view estimator. It computes angular differences between key bone vectors, and leverages a view encoder to extract spatial-temporal relationship for view prediction. 
}
    \label{fig:view estimator Principle}
\end{figure}

\begin{figure*}
\centering
     \setlength{\abovecaptionskip}{0.cm}
    \setlength{\belowcaptionskip}{0.cm}
\includegraphics[width=1\linewidth]{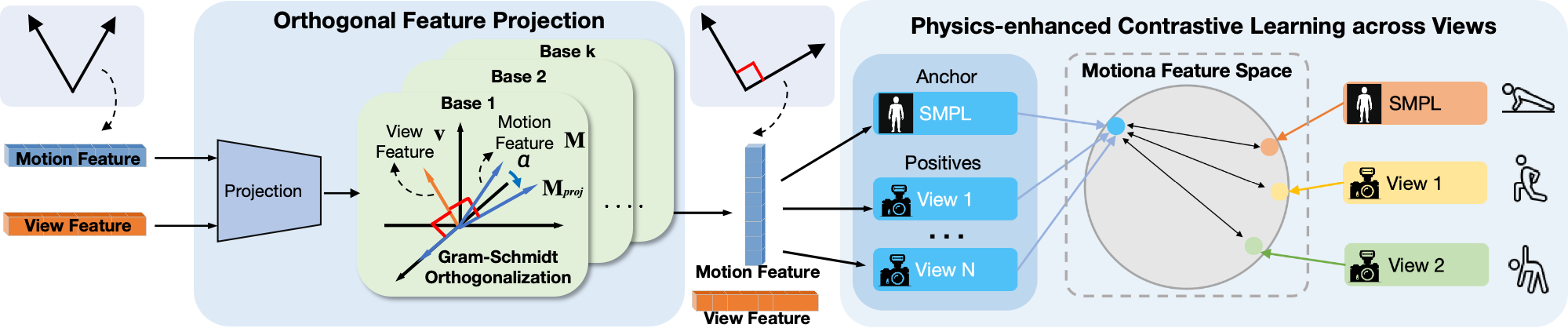}
\caption{View-orthogonal contrastive learning. We first disentangles motion and view features through the orthogonal projection module, and then extracts view-invariant motion features through physics-enhanced contrastive learning across different views. 
}
\label{fig:Orthogonal Loss Principle}
\end{figure*}

Specifically, for each frame, we compute  a compact geometric feature that characterizes the body's lateral span and depth configuration: $\mathbf{f}_{\text{view}} = 
[\mathbf{v}_{\text{hip}},\;
 \mathbf{v}_{\text{shoulder}},\;
 z_{\text{hip}_L},\;
 z_{\text{hip}_R},\;
 z_{\text{shoulder}_L},\;
 z_{\text{shoulder}_R}], $
where $\mathbf{v}_{\text{hip}} = \mathbf{k}_{\text{hip}_L} - \mathbf{k}_{\text{hip}_R}$ 
and $\mathbf{v}_{\text{shoulder}} = \mathbf{k}_{\text{shoulder}_L} - \mathbf{k}_{\text{shoulder}_R}$
represent the hip and shoulder width vectors, respectively, 
while the four $z$-coordinates capture the relative depth of these key joints.
This feature efficiently encodes body orientation and depth information with minimal computational overhead.
Then we collect a sequence of these geometric features across $T$ frames $\mathbf{F}_{\text{view}} = \{\mathbf{f}_{\text{view},1}, \mathbf{f}_{\text{view},2}, ..., \mathbf{f}_{\text{view},T}\} $, where angular variations across frames capture viewpoint dynamics.

\subsubsection{View embedding extraction.} To transform geometric features into robust viewpoint embeddings, we employ a lightweight view encoder to capture spatial-temporal relationships across frames. Specifically, the geometric features $\mathbf{F}_{\text{view}} $ is passed through a lightweight multi-layer perceptron to obtain the view embeddings:
\[
\mathbf{V} = {Enc}_{\text{view}}(\mathbf{F}_{\text{view}}) \in \mathbb{R}^{T \times D_{view}}.
\]
The view encoder consists of four fully connected layers, each followed by Layer Normalization, GELU activation, and Dropout.
This lightweight yet effective architecture captures essential spatial information correlated with camera viewpoint, generating view features $\mathbf{V} \in \mathbb{R}^{T \times D_{view}}$ that serve as compact, viewpoint-aware representations to enhance the robustness of pose estimation.

As a result, the view embedding compactly encodes viewpoint information by integrating both spatial geometrics and temporal dynamics. It is derived from intermediate pose features during inference, thus incurring minimal computational overhead and efficient for real-time use. The view embedding serves two key purposes: It guides view-invariant pose decoding (Section \ref{Subsection:View-Invariant_pose_estimation}), helping the model maintain accuracy across diverse camera angles. It identif ies challenging viewpoints that trigger the flip refinement module (Section \ref{subsection:View-Aware_adaptive_Inference}), which adaptively optimizes inference latency and improves pose estimation under ambiguous conditions. This design improves both the robustness of pose estimation and the efficiency of runtime inference.

\subsection{View-Invariant Pose Estimation}
\label{Subsection:View-Invariant_pose_estimation}

We aim to learn view-invariant motion representations through two steps: 1) an orthogonal projection module that disentangles motion features from view information, and 2) a physics-enhanced contrastive learning approach that enforces consistency of motion representations across different viewpoints.

\subsubsection{Disentangling motion and view features through orthogonal feature projection} 
To eliminate view-dependent biases from motion representations, we first project motion features into a subspace orthogonal to the view embeddings. 
 Given the initial motion and view embeddings $\mathbf{M}_{init} \in \mathbb{R}^{T \times D_{motion}}$ and $\mathbf{V} \in \mathbb{R}^{T \times D_{view}}$, we perform orthogonal projection:
$$\mathbf{M}_{proj} = \text{OrthoProjection}(\mathbf{M}_{init}, \mathbf{V}).$$

Specifically, to model complex view-dependent variations, we first generate multiple orthogonal bases from the view features $\mathbf{V}$ using a learned basis generator. For each frame $t$, the generator produces $K$ basis vectors $\mathbf{v}_k \in \mathbb{R}^{D{\text{base}}}$ ($k = 1,\dots,K$), each representing a principal direction along which motion features correlate with viewpoint. 
Multiple bases enable the model to capture diverse view-dependent effects across camera angles. Projecting motion features onto the orthogonal complement of these bases ensures robust disentanglement of viewpoint information.

Then we apply Gram-Schmidt orthogonalization process~\cite{hazewinkel2001orthogonalization} to remove view-dependent components from motion features by enforcing orthogonality between feature subspaces. For each basis vector $\mathbf{v}_k \in \mathbb{R}^{T \times D_{base}}$ (where $k = 1, ..., K$), we update:
$$\mathbf{M}_{proj} \leftarrow \mathbf{M}_{proj} - \alpha_k \cdot \mathbf{v}_k,$$
$$\alpha_k = \frac{\langle\mathbf{M}_{proj}, \mathbf{v}_k\rangle}{\|\mathbf{v}_k\|^2 + \epsilon}.$$
Here $\alpha_k$ is the projection coefficient indicating the component of $\mathbf{M}{\text{proj}}$ aligned with $\mathbf{v}k$. Subtracting this term removes view-aligned information so that the resulting motion feature $\mathbf{M}{\text{proj}}$ lie in the orthogonal complement of the motion base vector. And we set $\epsilon = 1e^{-6}$ to ensure numerical stability.  This procedure is applied sequentially over all $K$ bases, effectively removing view-aligned information from the motion features. 
After the orthogonal feature projection, we obtain motion features $\mathbf{M}_{ortho} \in \mathbb{R}^{T \times D_{motion}}$ that preserve temporal motion dynamics while being orthogonal to different view subspaces. 

We further enhance the disentanglement by introducing an orthogonality loss:
$$\mathcal{L}_{\text{ortho}} = \mathbb{E}[(\mathbf{M}_{proj} \cdot \mathbf{V})^2],$$
where the inner product $\mathbf{M}_{proj} \cdot \mathbf{V}$ measures the alignment between motion and view features, and squaring this inner product penalizes any remaining correlation between the two feature spaces.
The view estimator is trained without discrete viewpoint supervision: $\mathcal{L}_{\text{ortho}}$ enforces that the view features must capture all view-dependent information (otherwise view information leaks into motion features and increases the loss), while $\mathcal{L}_{\text{align}}$ enforces cross-view consistency via SMPL-based anchors; together, they provide continuous gradient signals for viewpoint learning. The discrete viewpoint labels (e.g., 10 camera IDs in HuMMan) are used solely for post-hoc validation in Section~\ref{sec:evaluation}, not for training.

\subsubsection{View-invariant motion feature extraction}
To enhance the consistency of motion representations across multiple viewpoints, we introduce a \emph{physics-enhanced contrastive alignment} module that enforces cross-view feature correspondence between SMPL-based physics dynamics and motion embeddings.

SMPL (Skinned Multi-Person Linear Model)~\cite{SMPL} represents the human body as a differentiable function of shape parameters $\boldsymbol{\beta} \in \mathbb{R}^{10}$ and pose parameters $\boldsymbol{\theta} \in \mathbb{R}^{72}$. A skinning function maps these parameters to a mesh of 6,890 vertices via linear blend skinning. Since SMPL encodes joint rotations in a body-centered coordinate system, its pose parameters are inherently view-invariant, providing view-independent supervision for our training.

The SMPL pose parameters \cite{SMPL} encode joint rotations in a canonical body coordinate system, making them inherently view-invariant and an effective anchor for enforcing consistent motion representations across viewpoints. Therefore, we use ground-truth SMPL parameters during training as supervision for physics-consistent dynamics.

Specifically, given paired SMPL and motion features from the same frame, we first project the original SMPL parameters into new embeddings $\mathbf{z}_{SMPL}$ to match the dimensionality of motion embeddings $\mathbf{z}_M$. Then we apply contrastive learning across motion features from different views, using the SMPL embedding as the positive anchor. This physics-enhanced contrastive learning pulls each motion embedding toward its corresponding SMPL embedding while pushing it away from non-matching instances, guiding motion representations toward a canonical, view-invariant space.

Given a batch of $N$ paired pose and motion embeddings, we concatenate both modalities to form a $2N$-sized representation set and compute:
\begin{equation}
\mathcal{L}_{\text{align}} = -\frac{1}{2N}\sum_{i=1}^{2N}\log \frac{\exp(\text{sim}(\mathbf{z}_i, \mathbf{z}_p) / \tau)}{\sum_{j=1}^{2N} \mathbb{1}_{[j \neq i]} \exp(\text{sim}(\mathbf{z}_i, \mathbf{z}_j) / \tau)},
\end{equation}
where $\mathbf{z}_i$ represents the L2-normalized embedding($\mathbf{z}_{SMPL}$ or $\mathbf{z}_M$), $\text{sim}(\cdot, \cdot)$ denotes cosine similarity, and $\tau$ is the temperature parameter controlling the sharpness of the distribution. The positive pair index $p(i)$ is defined as: $p(i) = i + N$ for $i \in [1, N]$ (pose$\rightarrow$motion), and $p(i) = i - N$ for $i \in [N+1, 2N]$ (motion$\rightarrow$pose). Self-contrast terms on the diagonal are masked to avoid trivial solutions. This bidirectional alignment ensures that pose and motion features from the same action instance maintain high similarity regardless of viewpoint variations.

Therefore, the overall training loss function can be define as:
\begin{equation}
\mathcal{L}_{\text{total}} = \alpha\mathcal{L}_{\text{ortho}} + \beta\mathcal{L}_{\text{align}},
\end{equation}
where $\alpha$ and $\beta$ balance motion-view feature disentanglement and motion feature alignment across views. This  ensures that the motion features generated by the model are both robust to viewpoint changes and discriminative for downstream tasks.

\subsection{View-Aware Adaptive Inference} 
\label{subsection:View-Aware_adaptive_Inference}

\begin{figure}
    \centering
         \setlength{\abovecaptionskip}{0.cm}
    \setlength{\belowcaptionskip}{0.cm}
    \includegraphics[width=1\linewidth]{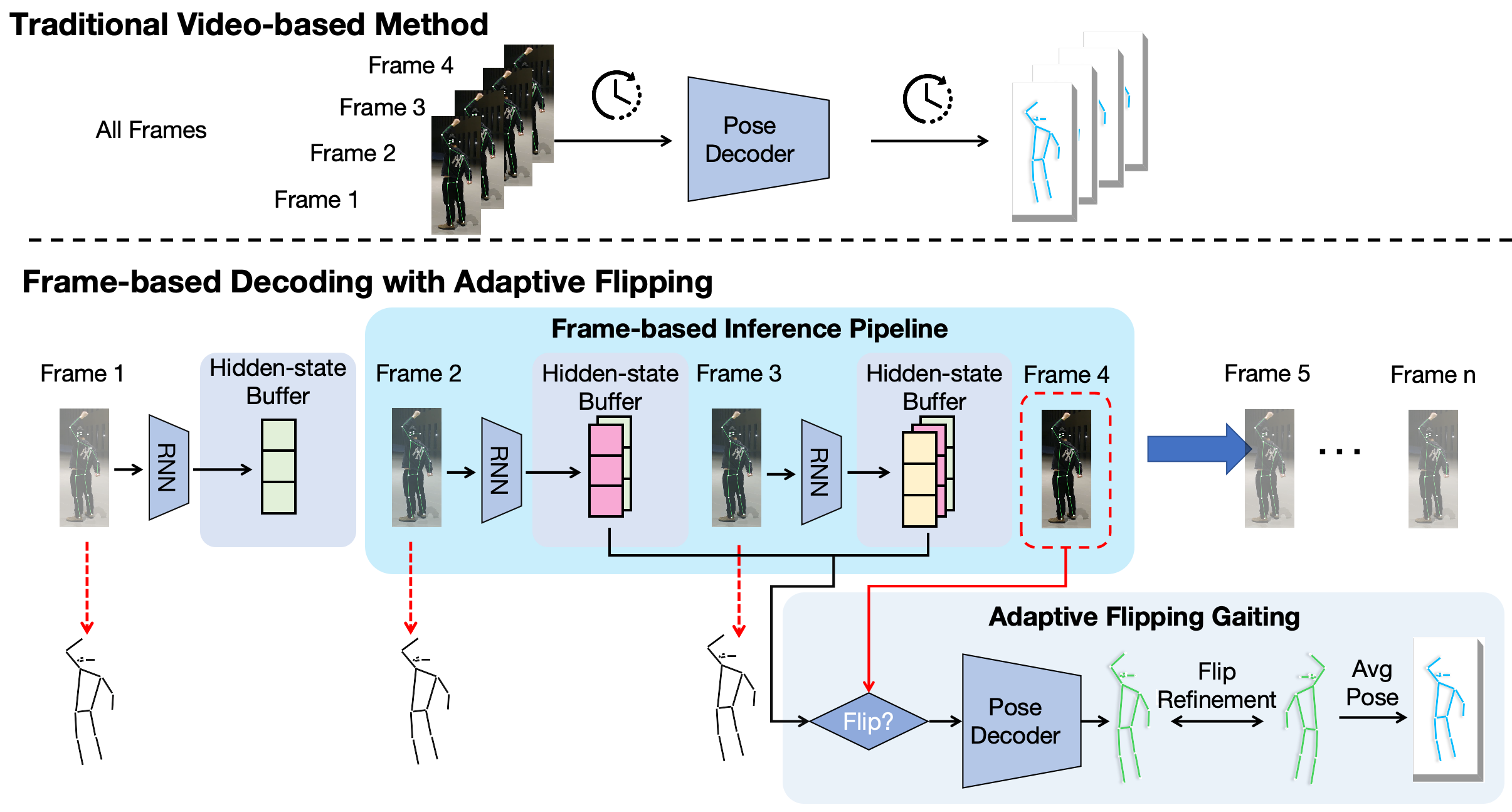}
    \caption{\name features a new frame-by-frame inference pipeline with adaptive activation of the flip refinement module to reduce inference latency. 
    }
    \label{fig:Inference}
\end{figure}

In this section, we propose to leverage viewpoint information to reduce inference latency of human pose estimation. As shown in Figure \ref{fig:Inference}, traditional pose estimation methods process entire video sequences, introducing significant inference latency from waiting for video streams. Moreover, it uniformly applies computationally expensive flipping refinement modules (e.g., 154.7 ms per frame) to all frames regardless of viewpoint complexity.
To enable real-time inference on edge devices, we design a frame-by-frame inference pipeline with adaptive activation of a flip refinement module. 

\subsubsection{Frame-by-frame inference pipeline} To enable real-time pose estimation with low latency, \name features a frame-by-frame inference pipeline that processes each frame as it arrives, supporting streaming prediction rather than full-sequence processing.

\textbf{Stateful temporal modeling with circular buffer.} 
To maintain temporal consistency during streaming inference, the decoder preserves recurrent states and previous outputs across frames, allowing incremental updates without recomputing the entire sequence. We implement a circular buffer that stores recurrent states and 3D keypoints from the most recent $W$ frames. This provides lossless short-term temporal context for stable reconstruction while ensuring constant memory usage. As new frames arrive, their features overwrite the oldest entries in constant time, maintaining an up-to-date window.

Compared to traditional framework based on the entire video sequence, our approach avoids redundant sequence processing and achieves constant per-frame latency. The combination of persistent states and windowed temporal features enables accurate, temporally coherent pose estimation in a truly online manner, making real-time 3D inference feasible even on resource-constrained edge devices.

\subsubsection{View-aware adaptive refinement}

While the streaming architecture achieves low latency, certain viewpoints (e.g., extreme angles, self-occlusions) remain challenging for pose estimation. To address this without sacrificing efficiency, we introduce a \emph{selective flip refinement} mechanism guided by viewpoint analysis.

Recognizing that the flipping refinement module is computationally intensive in pose estimation, we propose to selectively activate this module only for frames having suboptimal or challenging viewpoints.
Specifically, the flip refinement involves horizontally flipping the input, running inference on both versions, and averaging the resulting predictions to reduce left--right ambiguities---a standard technique in pose estimation~\cite{Sun2019HRNet} that improves accuracy but doubles computational cost. The hidden states from the original (unflipped) sequence are preserved and used for temporal bridging even when flip refinement is activated, maintaining motion continuity.
To identify such frames, \name utilizes viewpoint information extracted from the view estimator to assess whether the frames exhibit challenging viewing conditions, such as views learned from experiments that demonstrate tendencies toward degraded performance, that are likely to compromise pose accuracy. For these critical frames, the flip refinement module is activated to correct viewpoint-induced ambiguities and enhance the reliability of the pose predictions.
By confining the high-cost refinement process to only those frames where it is most beneficial, the system maintains real-time performance without sacrificing precision in complex scenarios.

\section{Evaluation}
\label{sec:evaluation}

\subsection{Methodology}

\subsubsection{Datasets}
Table~\ref{datasets} summarizes the datasets used in our evaluation.
\name is trained on Human3.6M~\cite{Ionescu2014Human36MLS}, MPI-INF-3DHP~\cite{mono-3dhp2017}, and InstaVariety~\cite{InstaVariety}, using SMPL parameters, 3D keypoints, and 2D keypoints as available, which exposes the model to both laboratory and in-the-wild pose variations. We evaluate on HuMMan~\cite{cai2022humman}, RICH~\cite{rich}, EMDB~\cite{kaufman2023emdb}, 3DPW~\cite{3dpw}, EgoHumans~\cite{khirodkar2023egohumans}, and two self-collected datasets for UAV tracking and gait analysis, covering indoor capture, in-the-wild scenes, moving cameras, fisheye distortion, and gait tasks.

\subsubsection{Evaluation Metrics}  
We use MPJPE, PA-MPJPE, PVE, and ACCEL to evaluate spatial accuracy, shape quality, and motion dynamics. MPJPE measures joint error directly, PA-MPJPE reduces rigid alignment effects, PVE evaluates mesh quality, and ACCEL reflects temporal smoothness. We also report ACC and F1-score.

\subsubsection{Baselines.}
We compare against representative 3D human pose and shape estimation baselines, including HMR2.0 \cite{goel2023humans}, ReFit \cite{wang23refit}, HSMR \cite{xia2025reconstructinghumansbiomechanicallyaccurate}, and WHAM \cite{shin2023wham}. For robustness analysis, we further compare with SPIN \cite{Kolotouros2019SPIN}, GraphCMR \cite{Kolotouros2019GraphCMR}, and the viewpoint-aware method NToP \cite{NToP}. All baselines are trained on the same datasets (Human3.6M, MPI-INF-3DHP, InstaVariety) with their official configurations for fair comparison.

\begin{table}
\centering
\small
\resizebox{\linewidth}{!}{
\begin{tabular}{lccc}
\toprule
\textbf{Dataset} & \textbf{Views} & \textbf{Modalities} & \textbf{Usage} \\
\midrule
Human3.6M~\cite{Ionescu2014Human36MLS} & 4 & RGB, 3D Keypoints, SMPL & Training \\
MPI-INF-3DHP~\cite{mono-3dhp2017} & 14 & RGB, 3D Keypoints, SMPL & Training \\
InstaVariety~\cite{InstaVariety} & Varied & RGB, 2D Keypoints & Training \\
AMASS~\cite{AMASS:2019} & Synthetic & 3D Keypoints, SMPL & Training \\
\midrule
HuMMan~\cite{cai2022humman} & 10 & RGB, Depth, 3D Keypoints, SMPL & Testing \\
RICH~\cite{rich} & 8 & RGB, Depth, 3D Keypoints, Contact & Testing \\
EMDB~\cite{kaufman2023emdb} & 6 & RGB, 3D Keypoints, SMPL & Testing \\
3DPW~\cite{3dpw} & 3 (not synchronized) & RGB, 3D Keypoints, SMPL & Testing \\
EgoHumans~\cite{khirodkar2023egohumans} & 4 (fisheye) & RGB, 3D Keypoints, SMPL & Testing \\
\parbox[t]{3.5cm}{UAV Dataset \textbf{(self-collected)}} & Continous Change & RGB, 3D Keypoints & Testing\\
\parbox[t]{3.5cm}{Gait Analysis \textbf{(self-collected)}} & 4 & RGB, Depth, 3D Keypoints, Infrared & Testing\\

\bottomrule
\end{tabular}}
\caption{Summary of datasets used in our experiments. The model is trained on Human3.6M, MPI-INF-3DHP, InstaVariety, and AMASS, and evaluated on HuMMan, RICH, EMDB, 3DPW, EgoHumans and two self-collected datasets.}
\label{datasets}
\end{table}

\begin{table*}
\centering
\resizebox{\textwidth}{!}{%
\begin{tabular}{c c ccc ccc ccc}
\toprule
\multicolumn{1}{c}{\textbf{Category}} & \multicolumn{1}{c}{\textbf{Approach}} & 
\multicolumn{3}{c}{\textbf{HuMMan Dataset (14 joints)}} &
\multicolumn{3}{c}{\textbf{EMDB Dataset (24 joints)}} &
\multicolumn{3}{c}{\textbf{RICH Dataset (24 joints)}} \\
\cmidrule(lr){3-5} \cmidrule(lr){6-8} \cmidrule(lr){9-11}
& & PA-MPJPE & MPJPE & PVE & PA-MPJPE & MPJPE & PVE & PA-MPJPE & MPJPE & PVE \\
\midrule

\multirow{4}{*}{\begin{tabular}{@{}l@{}}SOTA Pose \\Estimation \\Baselines\end{tabular}} 
& HMR2.0~\cite{goel2023humans}  & 78.5 & 115.1 & 132.8 & 59.9 & 98.7 & 120.2 & 62.8 & 103.6 & 108.3 \\
& ReFit \cite{wang23refit} & 79.1 & 117.4 & 135.2 & 71.2 & 104.2 & 123.9 & 67.3 & 93.4 & 113.4 \\
& HSMR \cite{xia2025reconstructinghumansbiomechanicallyaccurate} & 65.1 & 98.2 & 119.3 & 52.5 & 92.3 & 108.5 & 57.4 & 101.2 & 109.2 \\
& WHAM \cite{shin2023wham} & 61.1 & 99.7 & 112.1 & 52.8 & 77.7 & 93.6 & 55.7 & 98.3 & 111.5 \\
\midrule

\multirow{3}{*}{\begin{tabular}{@{}l@{}}Viewpoint-\\Invariant \\Baselines\end{tabular}} 
& SPIN ~\cite{Kolotouros2019SPIN} & 82.6 & 122.5 & 141.7 & 78.3 & 110.9 & 112.6 & 77.3 & 113.2 & 143.4 \\
& GraphCMR ~\cite{Kolotouros2019GraphCMR} & 85.3 & 126.8 & 145.1 & 86.6 & 112.3 & 141.5 & 69.5 & 102.4 & 113.8 \\
& NToP \cite{NToP} & 69.5 & 96.9 & 99.2 & 74.5 & 104.3 & 113.6 & 78.3 & 113.5 & 123.2 \\
\midrule

Our Method & \textbf{\name (Ours)} & \textbf{49.2} & \textbf{80.1} & \textbf{96.7} & \textbf{46.7} & \textbf{71.0} & \textbf{87.2} & \textbf{51.6} & \textbf{93.4} & \textbf{107.8} \\
\bottomrule
\end{tabular}%
}
\caption{Overall performance comparison of \name and various baselines on HuMMan, EMDB, and RICH datasets. Metrics include PA-MPJPE (mm), MPJPE (mm), and PVE (mm). Lower values are better for all metrics ($\downarrow$). Best result is in \textbf{bold}. }
\label{tab:comprehensive_comparison_final_perfected}
\end{table*}

\begin{figure}
  \centering
      \setlength{\abovecaptionskip}{0.cm}
    \setlength{\belowcaptionskip}{-0.cm}
  \includegraphics[width=\columnwidth]{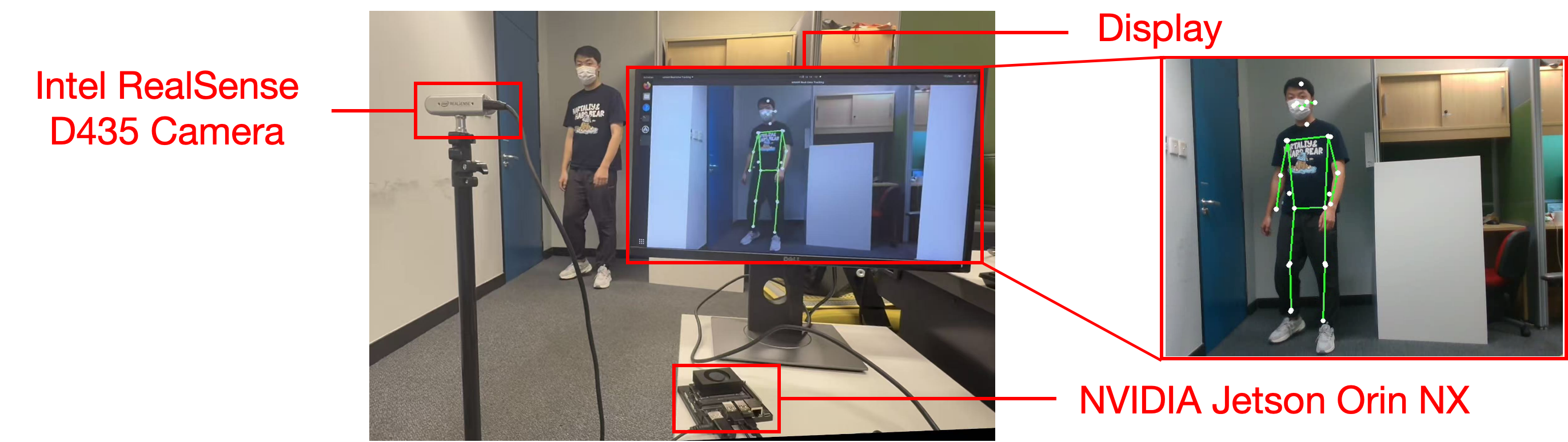}
  \caption{The end-to-end system for real-time pose estimation.}
  \label{fig:real-time}
\end{figure}

\subsubsection{Implementation Details.}

We use ViTPose~\cite{Xu2022ViTPoseSV} to extract 2D keypoints with a confidence threshold of 0.3. The motion encoder uses an RNN backbone, the view encoder is a four-layer MLP, and videos are processed in temporal chunks with normalized keypoints to preserve motion continuity. We train \name with Adam, an initial learning rate of $1{\times}10^{-4}$, MultiStep decay, batch size 64, and 40 epochs on $224{\times}224$ 16-frame clips.

Figure \ref{fig:real-time} shows the end-to-end setup, where Intel RealSense and fisheye cameras stream video to an Nvidia Jetson Orin NX for on-device inference. The RealSense D435 provides 30 FPS RGB frames at $1280\times720$, while the fisheye cameras provide 60 FPS at the same resolution.
This setup allows us to evaluate the same pipeline under both standard and wide-angle sensing conditions on an edge-class device.



\subsection{Overall Performance}

\subsubsection{Performance on different datasets.}
We benchmark \name against recent pose-estimation models and viewpoint-invariant baselines on HuMMan, EMDB, and RICH using the same training data for fair comparison.
\name achieves the best results on all three datasets: 49.2~mm PA-MPJPE on HuMMan, 46.7~mm on EMDB, and 51.6~mm on RICH. On EMDB and RICH, the gains show that MoViD remains effective under occlusion, human-object interaction, and varied camera placement. The largest gain appears on HuMMan, which has the most diverse and imbalanced viewpoint distribution, indicating that our method is especially effective under unseen or challenging views.
Compared with WHAM, the relative improvement on HuMMan is the most pronounced, which is consistent with our design goal of reducing view-dependent bias under large viewpoint variation. The consistent gains in MPJPE and PVE further indicate that MoViD improves not only aligned pose accuracy but also absolute joint localization and mesh reconstruction quality.
This cross-dataset pattern suggests that the benefit of explicit motion-view disentanglement is not tied to a single benchmark, but transfers across both controlled capture setups and in-the-wild scenes.

\begin{figure}
    \raggedright   
    \setlength{\abovecaptionskip}{0.cm}
    \setlength{\belowcaptionskip}{-0.cm}
    \begin{subfigure}{0.49\linewidth}
            \setlength{\abovecaptionskip}{0.cm}
        \setlength{\belowcaptionskip}{0.cm}
        \raggedright   
        \includegraphics[width=1\linewidth]{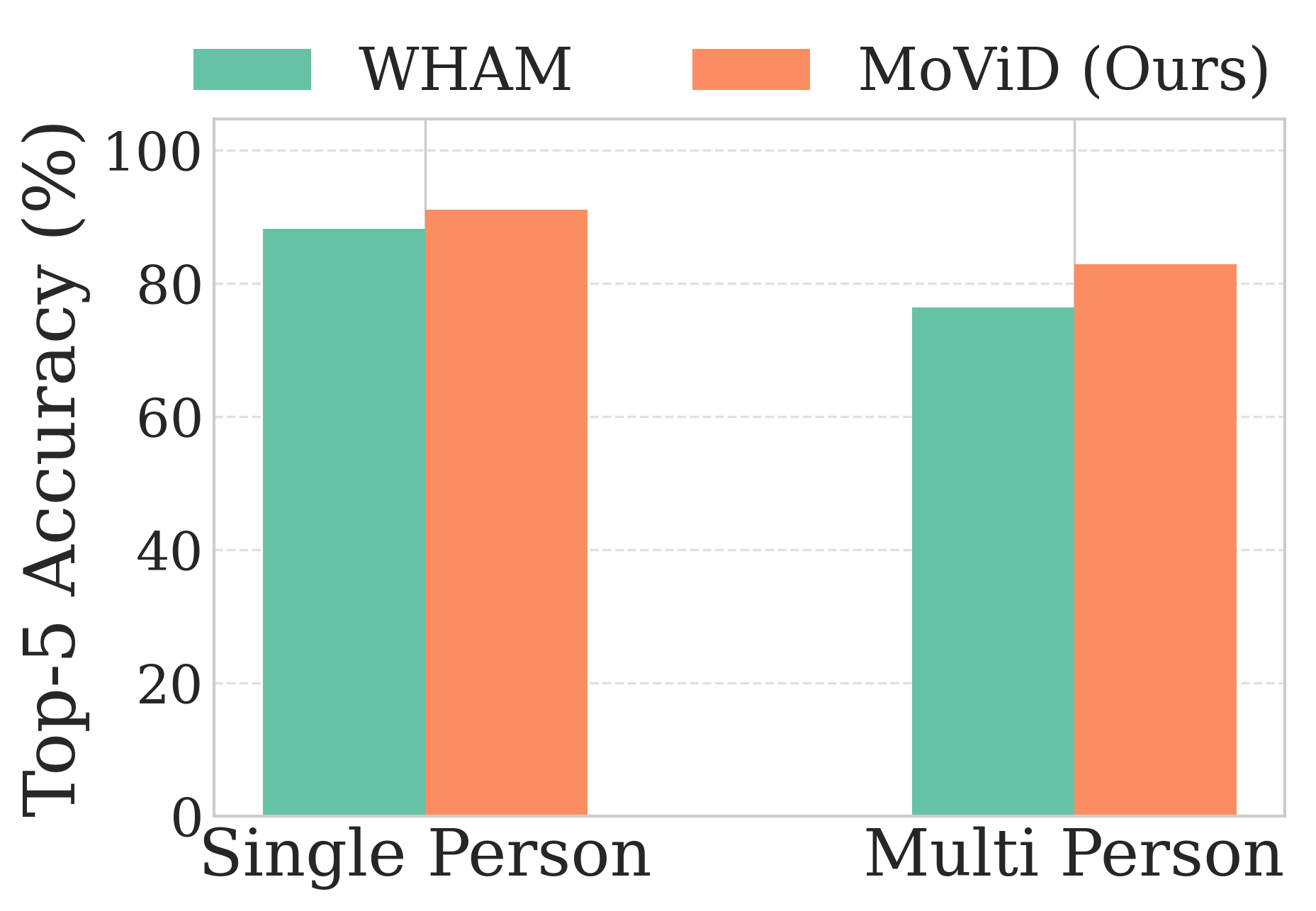}
        \caption{Action recognition}
        \label{fig:downstream_task_action_recognition}
    \end{subfigure}%
    \begin{subfigure}{0.49\linewidth}
        \centering
        \setlength{\abovecaptionskip}{0.cm}
        \setlength{\belowcaptionskip}{0.cm}
        \includegraphics[width=1\linewidth]{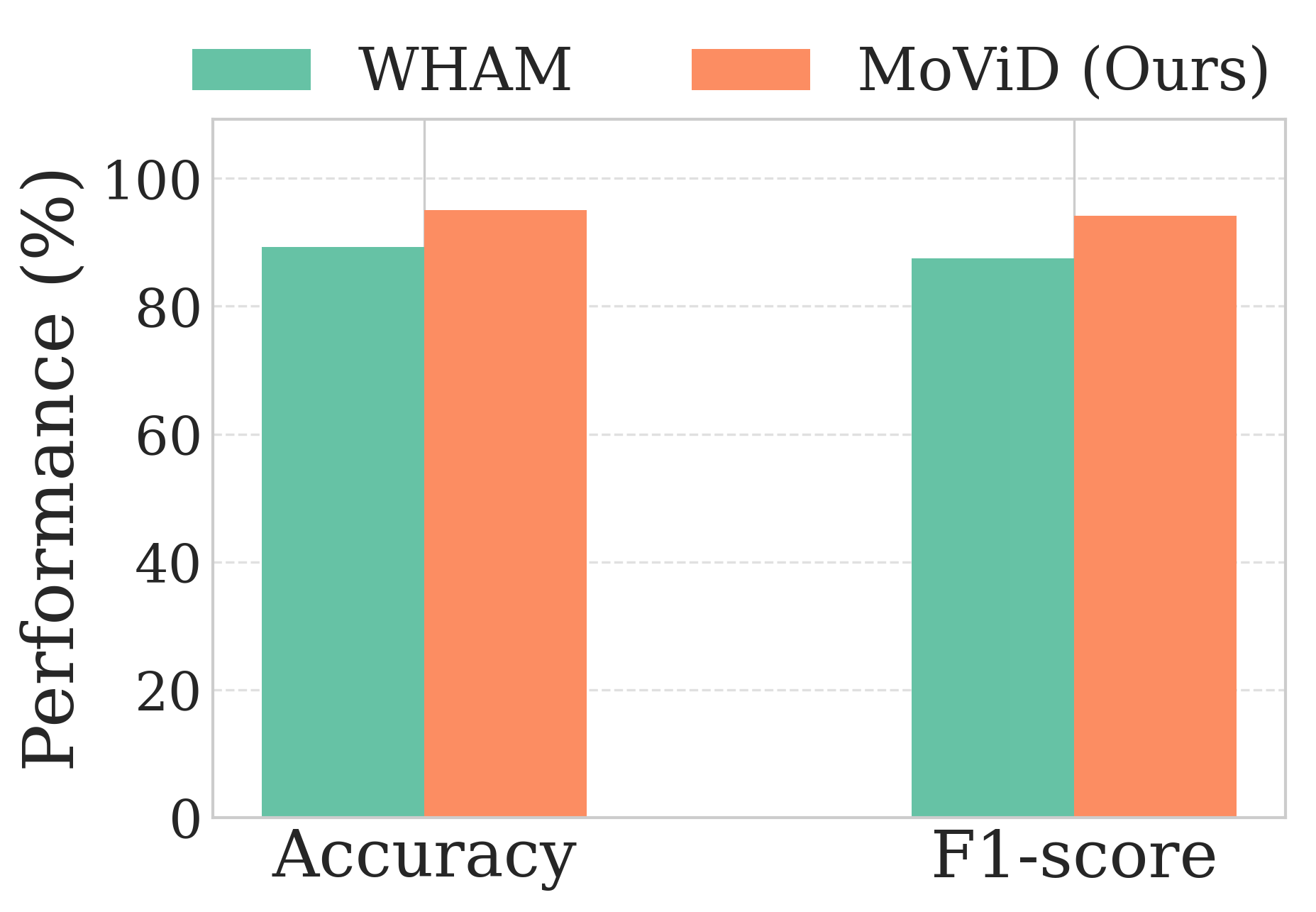}
        \caption{Fall detection.}
        \label{fig:downstream_task_fall_detection}
    \end{subfigure}
    \caption{Accuracy of downstream tasks.}
    \label{fig:downstream_task}
\end{figure}

\subsubsection{Downstream task performance}
\label{subsec:downstream}

We further evaluate \name on action recognition and fall detection. On NTU RGB+D \cite{liu2020ntu}, our viewpoint-invariant skeletons are used by ST-GCN and 2s-AGCN and yield competitive accuracy across five viewpoints. On UP-Fall \cite{s19091988}, \name reaches an F1-score of 94.1\%, showing that better cross-view skeleton quality also benefits downstream tasks.
These results suggest that the benefit of view-invariant pose estimation extends beyond upstream reconstruction metrics and translates to more reliable representations for recognition and safety-critical monitoring.
They also indicate that the learned skeleton representation can be reused effectively by standard downstream backbones without additional view-specific adaptation.


\begin{figure}
    \raggedright   
    \setlength{\abovecaptionskip}{0.cm}
    \setlength{\belowcaptionskip}{-0.cm}
    \begin{subfigure}{0.49\linewidth}
            \setlength{\abovecaptionskip}{0.cm}
        \setlength{\belowcaptionskip}{0.cm}
        \raggedright   
        \includegraphics[width=1\linewidth]{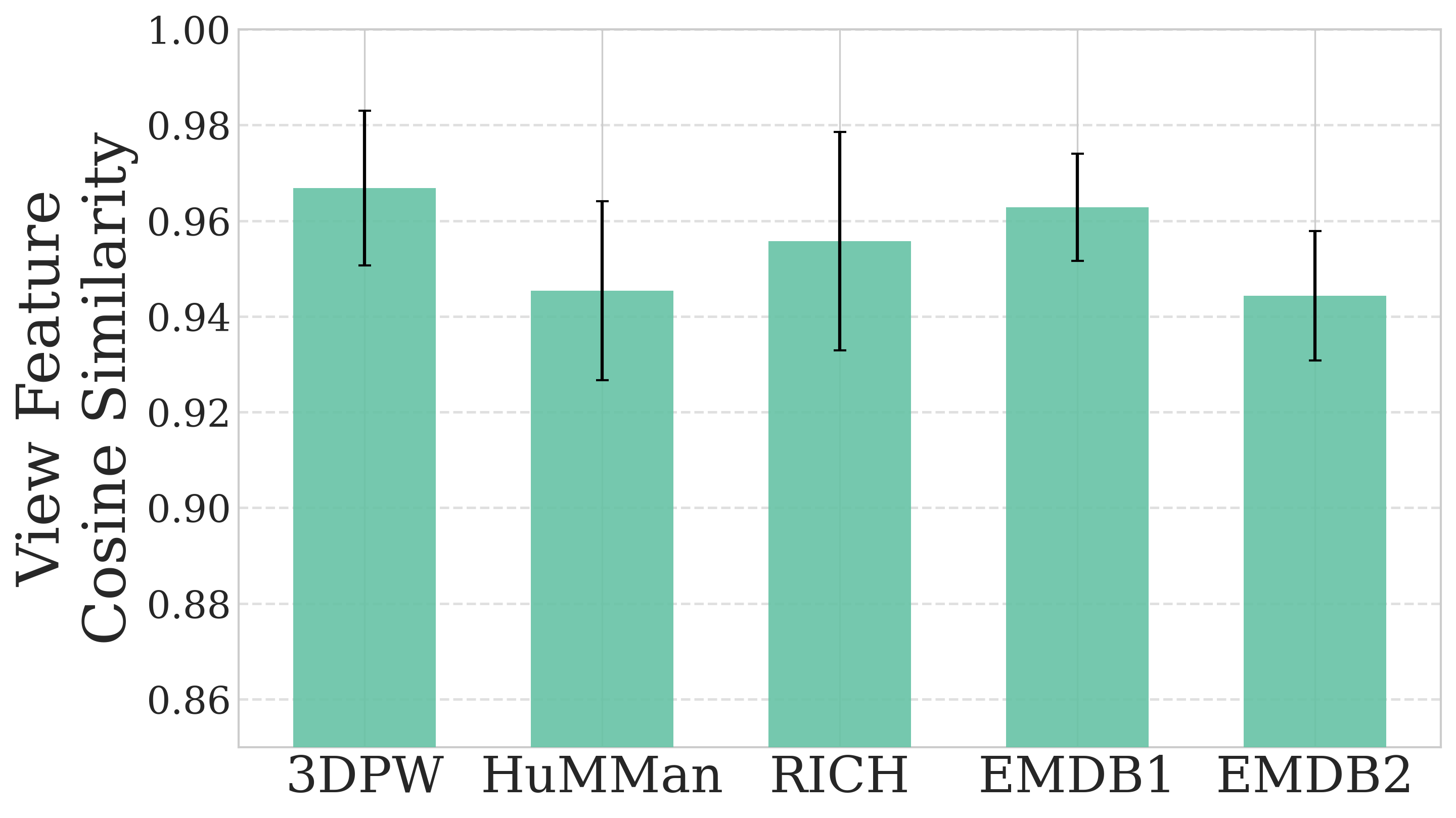}
        \caption{Viewpoint estimation performance across five datasets.}
        \label{fig:view_accuracy}
    \end{subfigure}%
    \begin{subfigure}{0.49\linewidth}
        \centering
        \setlength{\abovecaptionskip}{0.cm}
        \setlength{\belowcaptionskip}{0.cm}
        \includegraphics[width=1\linewidth]{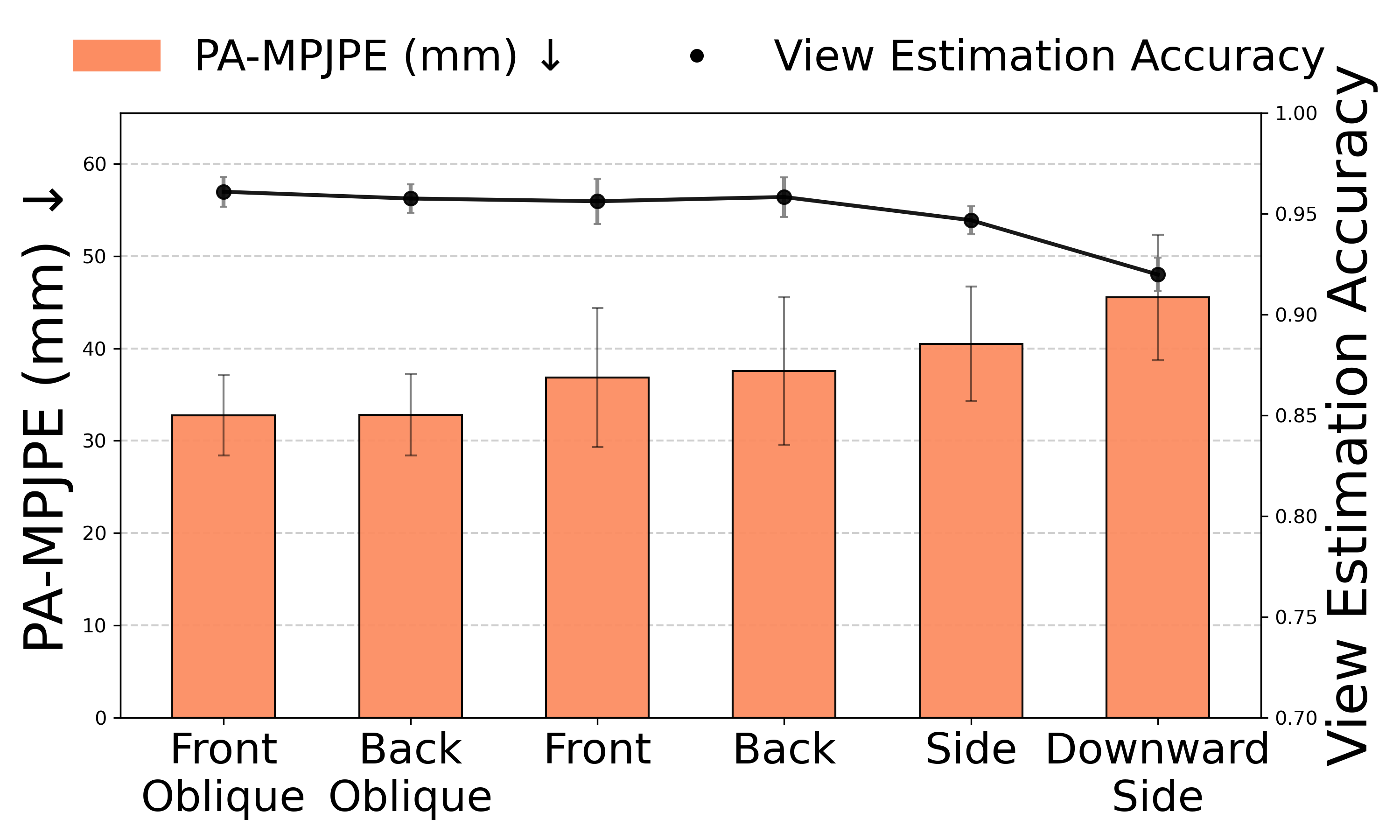}
        \caption{Inverse correlations with pose estimation accuracy.}
        \label{fig:similarity_vs_error}
    \end{subfigure}
    \caption{Viewpoint estimation performance and its relationship with pose estimation accuracy.}
    \label{fig:viewpoint_performance}
\end{figure}

\subsection{Impact of View Estimation}

\subsubsection{Viewpoint Estimation Accuracy.}
Accurate viewpoint estimation is central to MoViD. We evaluate the continuous view features by clustering them across 10 fixed camera views. As shown in Figure~\ref{fig:view_accuracy}, accuracy stays above 0.94 and peaks at 0.967, indicating that the learned features reliably capture viewpoint geometry across datasets.
The small variation across evaluation settings also indicates that the view estimator remains stable under differences in scene layout, subject appearance, and motion pattern.

\subsubsection{Impact on pose estimation errors.}

We further analyze the relationship between view estimation accuracy and pose error (PA-MPJPE) across six representative viewpoints (Figure~\ref{fig:similarity_vs_error}). Better view estimation consistently leads to lower pose error. The oblique front view performs best, likely because both lateral span and depth cues are visible, while the top-down side view remains the most difficult because of self-occlusion and compressed body geometry.
This trend supports the role of viewpoint estimation as a useful intermediate signal: when view features are more discriminative, the subsequent motion-view disentanglement becomes more effective and the final pose error decreases accordingly.





\subsubsection{Error Propagation Analysis.}
To assess robustness to upstream noise, we inject zero-mean Gaussian noise (0--250 mm) into the 2D and coarse 3D keypoints before refinement. MPJPE and PA-MPJPE increase only gradually, showing that MoViD suppresses accumulated errors under substantial perturbations.

\subsubsection{Comparison with other Viewpoint Estimation Methods.}
Compared with CNN-based regression~\cite{Ghezelghieh2016CNN3DV} and auxiliary viewpoint prediction~\cite{Wang2020PredictingCV}, \name achieves much higher view accuracy and lower downstream PA-MPJPE by explicitly enforcing view-invariant motion features (Table~\ref{tab:view_estimator_comparison}). This confirms that viewpoint prediction is most useful when it is directly tied to motion-view disentanglement.

\begin{table}[]
\centering
\small
\begin{tabular}{lcc}
\toprule
\textbf{Method} & \textbf{View Acc. (\%)} & \textbf{PA-MPJPE (mm)} \\
\midrule
CNN-based regression~\cite{Ghezelghieh2016CNN3DV} & 78.3 & 98.7 \\
Auxiliary prediction~\cite{Wang2020PredictingCV} & 80.1 & 84.3 \\
\textbf{\name (Ours)} & \textbf{98.7} & \textbf{49.2} \\
\bottomrule
\end{tabular}
\caption{Comparison of viewpoint estimation methods on HuMMan.}
\label{tab:view_estimator_comparison}
\vspace{-0.5cm}
\end{table}

\subsection{System Efficiency}
We evaluate system-level efficiency through overall performance comparison, runtime breakdown, and view-wise analysis.


\begin{table}
\centering
\small
\resizebox{8.5cm}{!}{
\begin{tabular}{lcccccccl}
\toprule
\textbf{Setting}  &  \textbf{Latency (ms)} &\textbf{Power Peak (W)}&  \textbf{PA-MPJPE}  & \textbf{ACCEL} & \textbf{Stream} \\
\midrule
WHAM           & 154.7 & 13.8 & 56.10  & 30.00 &  \XSolidBrush \\

\name       & \textbf{66.2} & \textbf{8.8} & \textbf{46.88}  & \textbf{27.51} &  \Checkmark \\
\bottomrule
\end{tabular}
}

\caption{Overall system performance comparison on the HuMMan dataset.}
\label{tab:overall_system_comparison}
\vspace{-0.5cm}
\end{table}

\begin{table}
\centering
\small
\resizebox{8.5cm}{!}{
\begin{tabular}{lcccc}
\toprule
\textbf{Module} & \textbf{2D Pose} & \textbf{Preprocessing} & \textbf{Inference}& \textbf{Total} \\
\midrule
\textbf{Latency (ms)} & 26.9 & 10.3 & 29.0 & \textbf{66.2} \\
\bottomrule
\end{tabular}
}
\caption{Runtime breakdown on Nvidia Jetson Orin NX.}
\label{tab:runtime_breakdown}
\vspace{-0.5cm}
\end{table}

\begin{table}[]
\centering
\small
\resizebox{8.5cm}{!}{
\begin{tabular}{lccc}
\toprule
\textbf{Scenario} & \textbf{\name Latency (ms)} & \textbf{End-to-End (ms)} & \textbf{FPS} \\
\midrule
Good view (flip off) & 25.0 & 62.2 & 16.1 \\
Bad view (flip on) & 62.9 & 100.1 & 10.0 \\
\bottomrule
\end{tabular}}
\caption{Latency under good and bad views on Nvidia Jetson Orin NX.}
\label{tab:latency_goodbad}
\vspace{-0.5cm}
\end{table}

\subsubsection{Overall system performance comparison}

We benchmark \name against WHAM on HuMMan in terms of both accuracy and efficiency. Table~\ref{tab:overall_system_comparison} shows that \name improves all accuracy metrics while reducing per-frame latency from 154.7 ms to 66.2 ms and peak power from 13.8 W to 8.8 W, enabling real-time streaming on edge devices. These gains are important for long-running mobile or embedded deployments where both latency and power budget matter.
This result shows that MoViD improves the usual accuracy-efficiency trade-off rather than exchanging one for the other, which is important for practical edge deployment.

\subsubsection{Latency of different design components}
\label{sec:runtime_analysis}
Table~\ref{tab:runtime_breakdown} reports a total latency of 66.2 ms per frame (15.1 FPS) on Nvidia Jetson Orin NX. Of this, ViTPose accounts for 26.9 ms, while preprocessing and MoViD inference together take 39.3 ms.
The breakdown shows that our additional viewpoint-aware modules introduce only moderate overhead relative to the upstream detector while still providing clear gains in pose accuracy.

\subsubsection{Latency under good and bad views}
Under good views, the flip refinement is skipped and latency is lower; under bad views, refinement is activated and latency increases (Table~\ref{tab:latency_goodbad}). ACCEL remains stable, showing that adaptive flipping does not introduce obvious temporal artifacts while still allocating more computation to harder viewpoints.

\begin{figure}
    \raggedright   
    \setlength{\abovecaptionskip}{0.cm}
    \setlength{\belowcaptionskip}{-0.cm}
    \centering
    \includegraphics[width=0.9\linewidth]{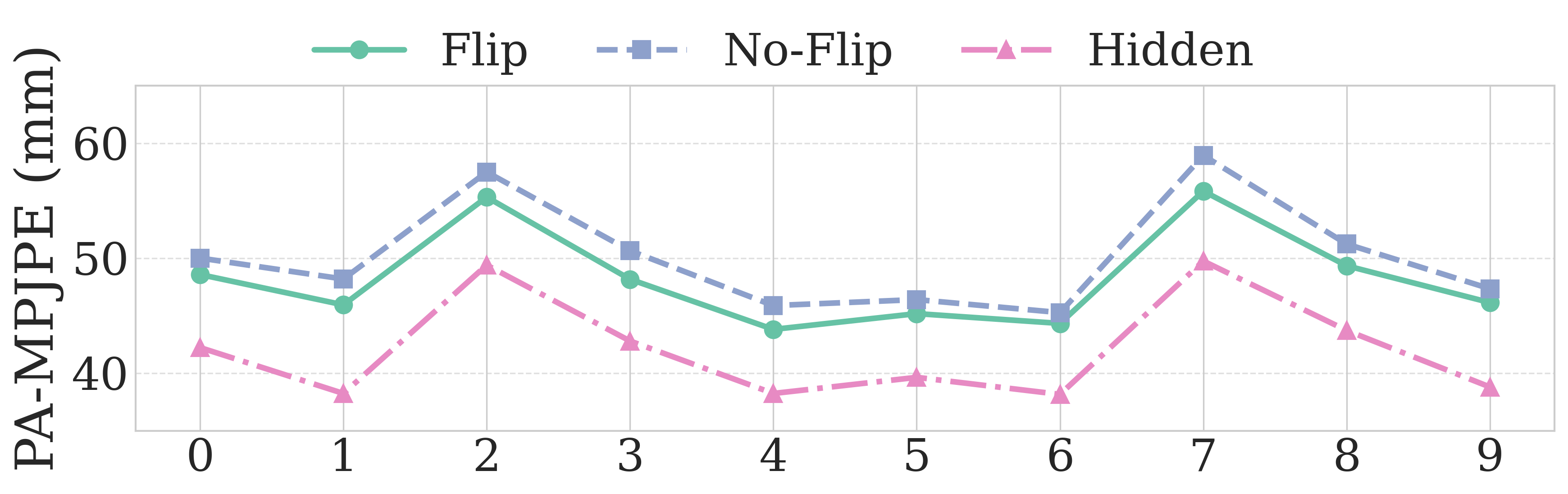}
    \caption{Effect of flip augmentation on different views.}
    \label{figure/Flip Augmentation}
\end{figure}

\subsubsection{Effectiveness on different views} 
We evaluate view-aware adaptive refinement across all 10 HuMMan views~\cite{cai2022humman}. As shown in Figure.~\ref{figure/Flip Augmentation}, flip augmentation reduces PA-MPJPE by 1.43 mm on average, especially on difficult side views. Temporal bridging improves motion continuity, indicating that the adaptive strategy improves difficult viewpoints without sacrificing stable temporal behavior.
The larger gains on side and downward views are consistent with our motivation that these perspectives suffer more from left--right ambiguity and partial self-occlusion, making selective refinement more beneficial there than on easier frontal views.

\subsection{Understanding MoViD performance}

\subsubsection{Ablation study} 
\begin{figure}
    \centering   
    \setlength{\abovecaptionskip}{0.cm}
    \setlength{\belowcaptionskip}{-0.cm}
        \begin{subfigure}{1\linewidth}
        \centering
        \setlength{\abovecaptionskip}{0.cm}
        \setlength{\belowcaptionskip}{0.cm}
        \includegraphics[width=\linewidth]{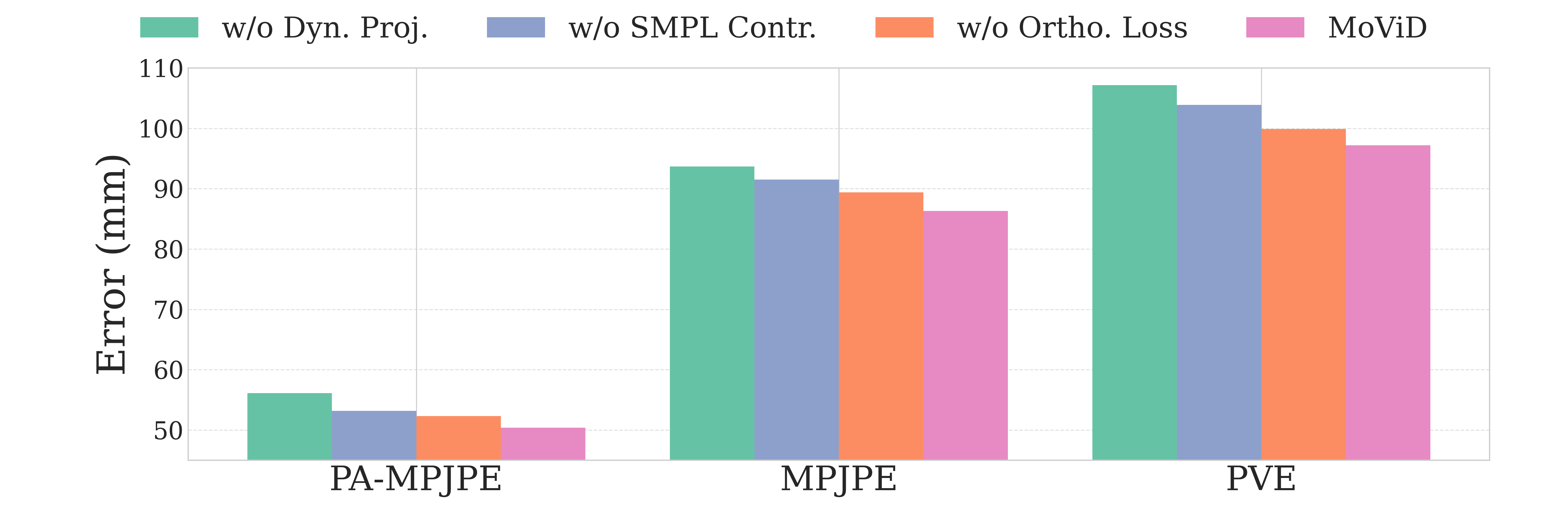}
        \caption{Ablation study.}
        \label{fig/ablation}
    \end{subfigure}
    \begin{subfigure}{0.48\linewidth}
            \setlength{\abovecaptionskip}{0.cm}
        \setlength{\belowcaptionskip}{0.cm}
        \centering   
        \includegraphics[width=1\linewidth]{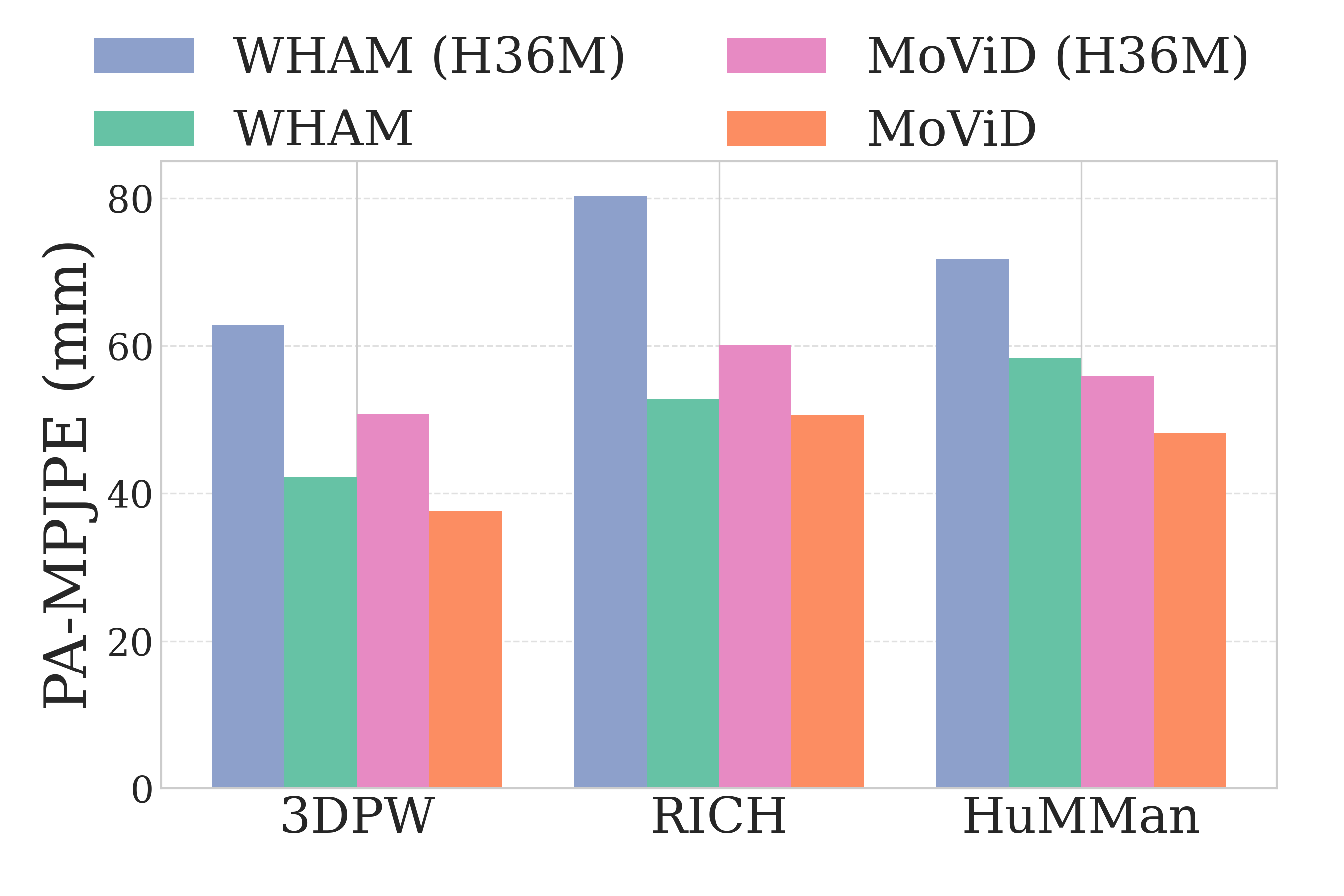}
        \caption{Limited training views.}
        \label{figure:pose_metrics_comparison}
    \end{subfigure}%
    \begin{subfigure}{0.5\linewidth}
        \centering
        \setlength{\abovecaptionskip}{0.cm}
        \setlength{\belowcaptionskip}{0.cm}
        \includegraphics[width=1\linewidth]{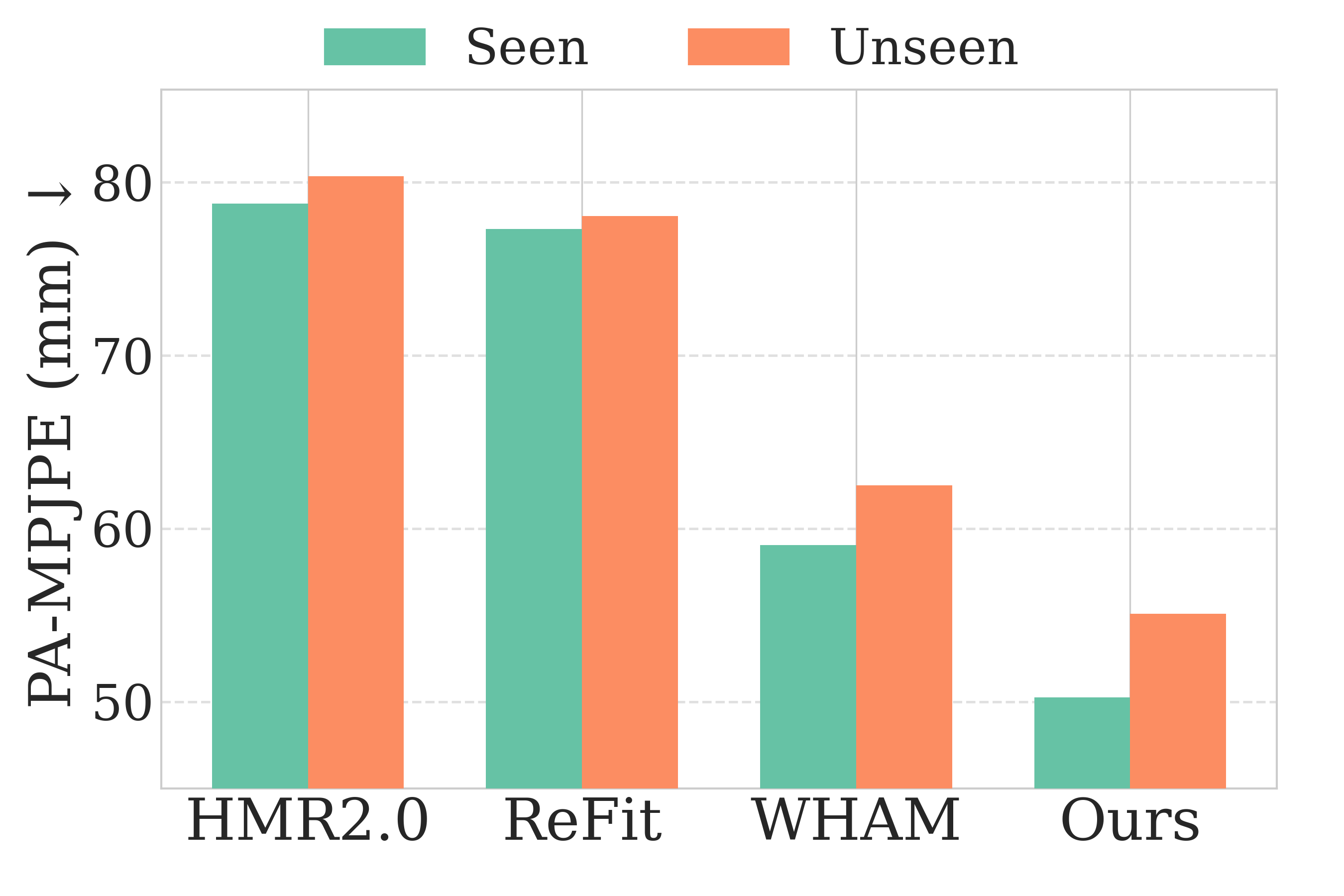}
        \caption{Seen and unseen views.}
        \label{figure:cross_view}
    \end{subfigure}%
    \caption{Understanding MoViD performance.}
    \label{fig:robustness_effectiveness}
\end{figure}

To assess the contribution of each core component in our framework, we conduct an ablation study on the HuMMan dataset, as shown in Figure~\ref{fig/ablation}. We evaluate the effect of removing three key modules: dynamic projection, orthogonal loss, and contrastive loss.
Dynamic projection is the most critical component: removing it increases PA-MPJPE from 50.4 mm to 56.1 mm. Removing contrastive loss raises PA-MPJPE to 53.2 mm, and removing orthogonal loss raises it to 52.3 mm. The full model performs best across all metrics, confirming that these components provide complementary benefits.
In particular, the large drop without dynamic projection shows that explicit motion-view separation is the main source of robustness, while orthogonal regularization and contrastive alignment further stabilize the learned representation.

\subsubsection{Performance when training with limited views.}
To test robustness under limited viewpoint diversity, we train only on Human3.6M~\cite{Ionescu2014Human36MLS}. Figure~\ref{figure:pose_metrics_comparison} shows that both our method and WHAM degrade, but \name generalizes much better, especially on RICH and HuMMan, where the test viewpoints differ substantially from the training set.
This result indicates that MoViD does not rely on dense viewpoint coverage during training to the same extent as conventional methods.

\subsubsection{Performance on unseen views}
We evaluate cross-view robustness on HuMMan~\cite{cai2022humman} using seen and unseen camera views. As shown in Table~\ref{figure:cross_view}, our method reaches 50.27 mm PA-MPJPE on seen views and 55.09 mm on unseen views, outperforming WHAM by 15.0\% and 12.0\%, respectively. This result suggests the learned motion representation transfers well even when the viewpoint was not observed during training.
The small gap between seen and unseen performance indicates that the disentangled motion features remain stable across viewpoint changes.

\subsection{Real-World Case Study}

In this section, we demonstrate the real-world applicability of \name by evaluating it on two new self-collected datasets: a dynamic UAV-based dataset and a controlled multi-modal gait dataset to enable comprehensive assessment of 3D pose estimation robustness under both fixed and moving camera conditions.

\subsubsection{UAV-Based Human Tracking Dataset}

While most existing datasets focus on controlled multi-view setups, real-world applications often involve cameras in motion, leading to continuous changes in viewpoint and perspective. To address this gap, we collected a UAV-based 3D human motion dataset that captures diverse human actions under dynamic camera trajectories.

\begin{figure}
    \raggedright   
    \setlength{\abovecaptionskip}{0.cm}
    \setlength{\belowcaptionskip}{-0.cm}
    \includegraphics[width=1\linewidth]{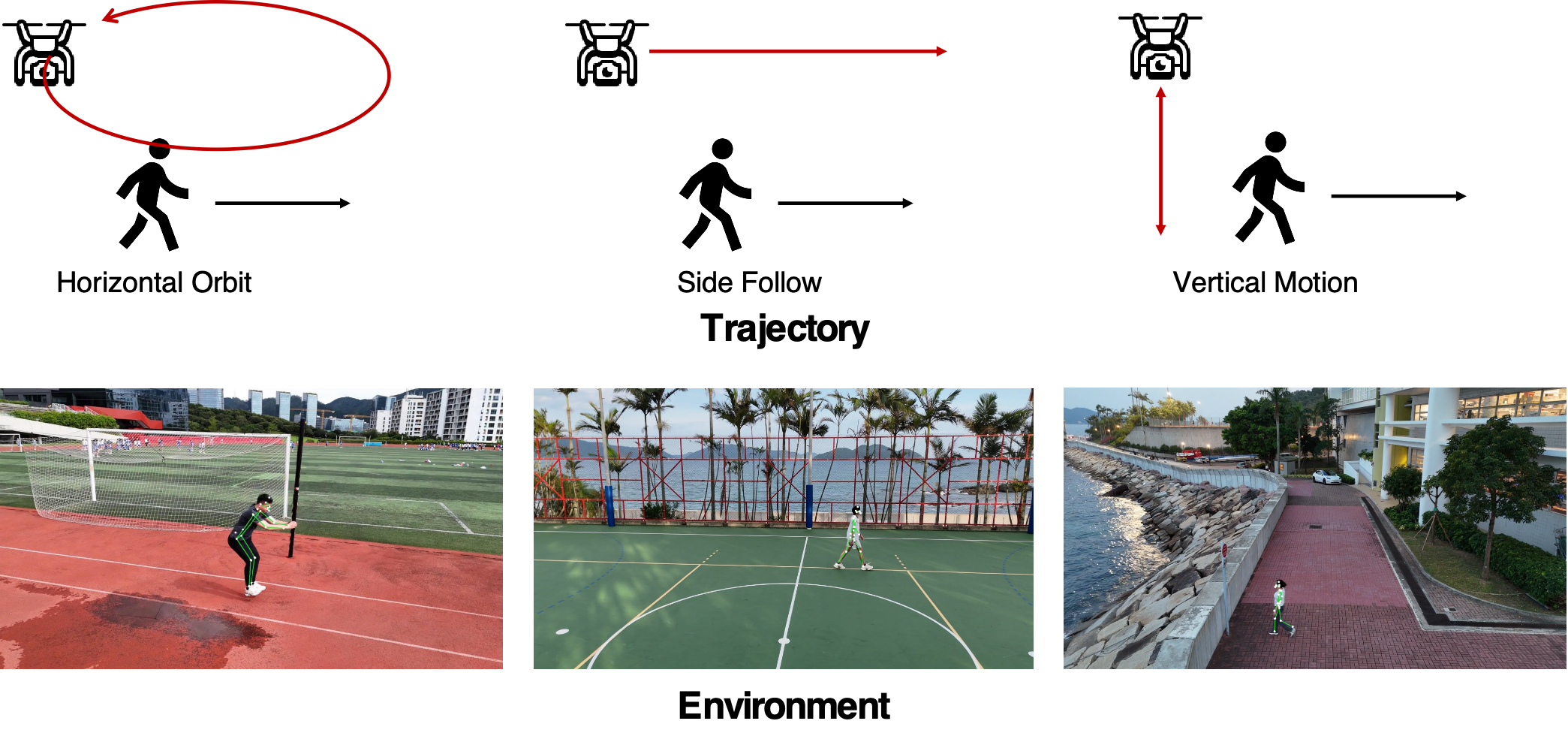}
    \caption{UAV trajectories including side-follow, horizontal orbit, and vertical motion, with subjects performing walking, jogging, hops, lunges, and mixed actions.}
    \label{fig:uav_trajectory_environmnet}
\end{figure}

\begin{figure}
    \raggedright   
    \setlength{\abovecaptionskip}{0.cm}
    \setlength{\belowcaptionskip}{-0.cm}
    \begin{subfigure}[t]{0.5\linewidth}
        \setlength{\abovecaptionskip}{0.cm}
        \setlength{\belowcaptionskip}{0.cm}
        \raggedright   
        \includegraphics[width=1\linewidth]{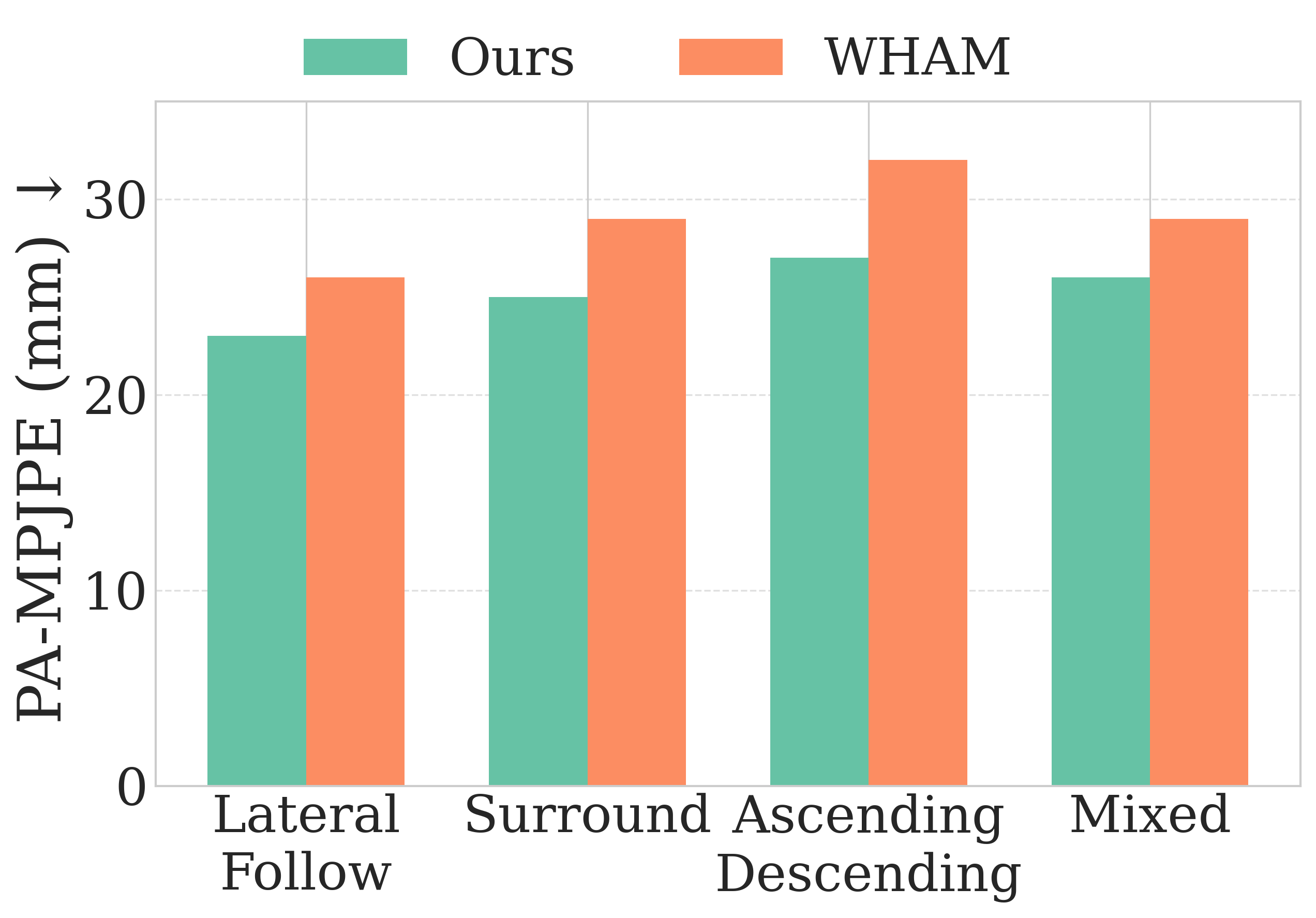}
        \caption{Mean estimation errors.}
        \label{fig:UAV_analysis_camera_motion}
    \end{subfigure}%
    \begin{subfigure}[t]{0.5\linewidth}
        \setlength{\abovecaptionskip}{0.cm}
        \setlength{\belowcaptionskip}{0.cm}
        \raggedright
        \includegraphics[width=1\linewidth]{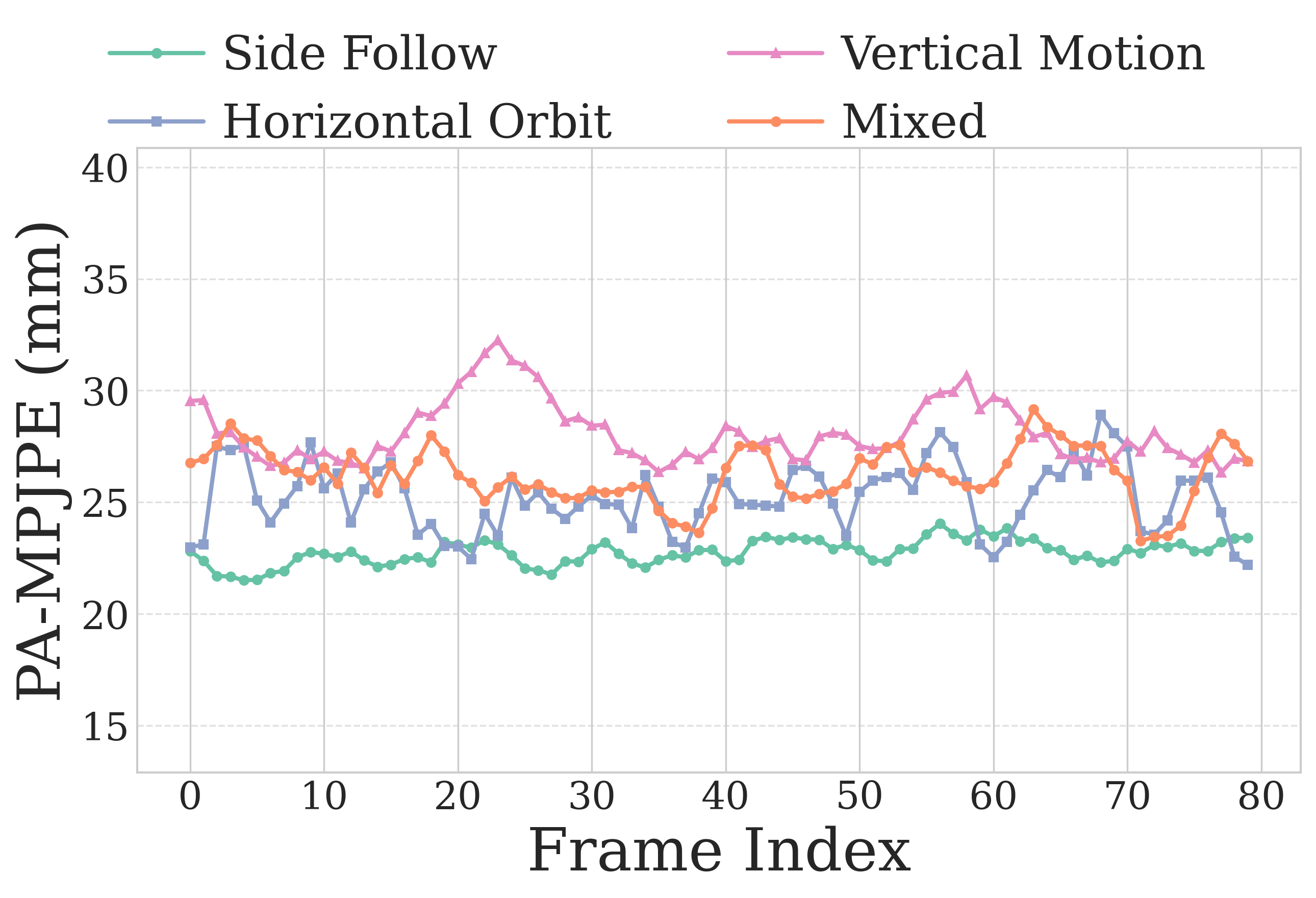}
        \caption{Errors over time.}
        \label{fig:uav_errors}
        \end{subfigure}
    \caption{Performance on the self-collected UAV dataset across different trajectories.}
    \label{figure:UAV_analysis}
\end{figure}

The dataset includes actions such as walking, jogging, single-leg hops, lunges, and mixed movement sequences, performed by 10 subjects of varying height and build. Each subject is recorded under four planned UAV trajectories: \textit{side-follow}, \textit{horizontal orbit}, \textit{vertical motion}, and \textit{mixed} trajectories that combine multiple movement patterns. These trajectories are designed to emulate realistic UAV tracking scenarios, including altitude changes, lateral shifts, and rotations around the subject.
To provide accurate ground truth for 3D human pose, a synchronized Intel RealSense D435 camera is placed on the ground for each recording. This provides calibrated 3D skeleton annotations with 17 keypoints per frame. The dataset comprises approximately 20 minutes of video per subject, captured at 30 FPS, resulting in over 360,000 frames in total. Recordings span outdoor environments with varied backgrounds and lighting conditions, introducing realistic challenges in occlusion, low light, and scale variation.
By combining continuous viewpoint changes, realistic UAV trajectories, and diverse human motions, this dataset is particularly suited for evaluating 3D human pose estimation and tracking models under dynamic, unconstrained camera conditions.

We evaluate our method and WHAM using PA-MPJPE under different camera motion categories (Fig.~\ref{figure:UAV_analysis}). Our method consistently achieves lower PA-MPJPE across all dynamic conditions, showing 11–16\% improvements under camera motion. These gains indicate that our approach effectively maintains pose accuracy despite continuous viewpoint shifts. This dataset establishes a strong benchmark for evaluating 3D human pose estimation with moving cameras, bridging the gap between controlled environments and real-world aerial scenarios. 
The error curves in Figure~\ref{fig:uav_errors} also show that MoViD remains more stable under sustained trajectory changes, especially during orbiting and altitude variation where viewpoint changes are rapid.





\subsubsection{Multimodal Gait Analysis Dataset.}
To assess performance in everyday human activities, we collected a multi-modal gait dataset for estimating pose estimation accuracy and gait parameter prediction across different views. Gait analysis plays a vital role in understanding human motion and detecting early signs of mobility impairments. Existing gait datasets frequently lack coverage from multiple viewpoints and are typically restricted to a single sensing modality. 
To address these issues, we collected a new multi-modal dataset involving 20 participants performing standardized gait activities, including walking, stair climbing, TUG, bending, and squatting. Data were captured using four synchronized Intel RealSense D435 cameras positioned at multiple angles (front, side, and back) to ensure viewpoint diversity. The dataset integrates seven distinct modalities—RGB, Depth, Infrared, IMU, bone vectors, 3D skeleton and text prompt. This dataset supports research on multimodal fusion and cross-view generalization for gait understanding.

Compared to WHAM, \name consistently achieves lower PA-MPJPE across all actions and views. For example, in the walking task, \name reduces pose error by 10.5\% (front view), 12.9\% (left), 14.9\% (right), and 15.8\% (back). These results highlight the model's robustness to viewpoint changes and motion types. In addition to pose estimation, \name also improves gait parameter predictions. For cadence, the mean absolute error is reduced by 25.0\% on average across views, while stride length error drops by an average of 13.1\%. Overall, \name's performance gains indicate its strong potential for real-world gait assessment and clinical applications.
The improvement is not limited to a single activity: MoViD remains consistently better across walking, stairs, TUG, bending, and squatting, which suggests that the method supports both periodic gait analysis and more diverse rehabilitation motions.
Taken together with the UAV results, these findings show that \name remains robust when viewpoint variation arises from either moving cameras or subject-centric activity changes in practical deployments.
\begin{figure}
  \centering
    \setlength{\abovecaptionskip}{0.cm}
    \setlength{\belowcaptionskip}{-0.cm}
  \includegraphics[width=\columnwidth]{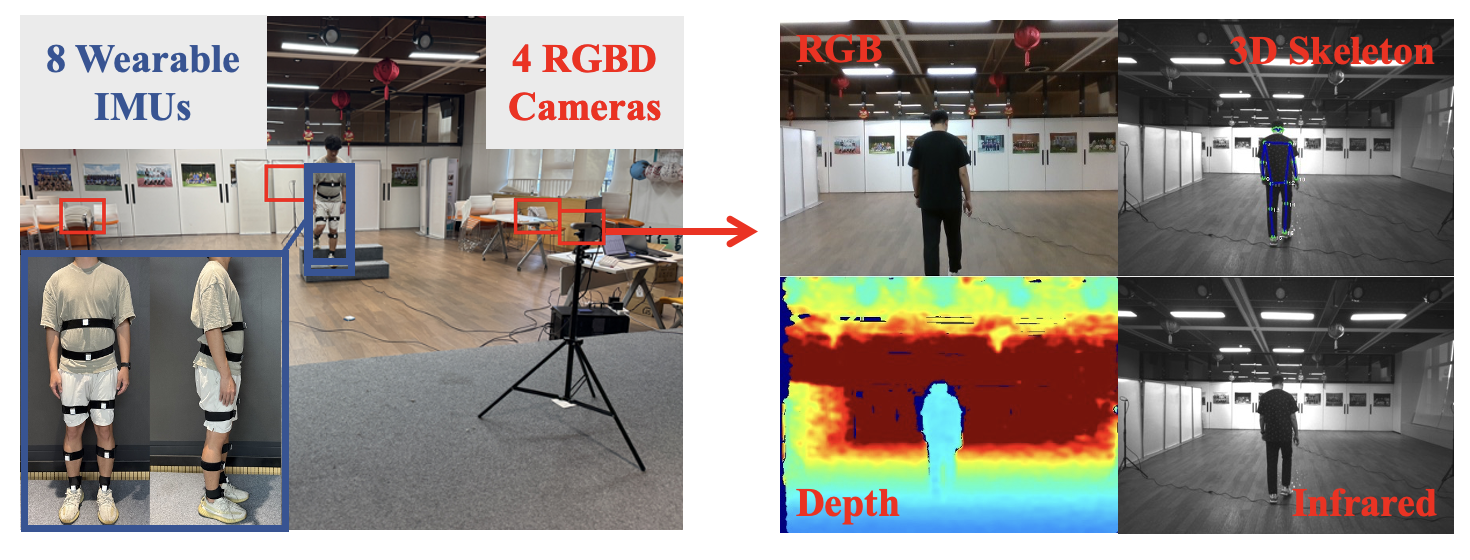}
  \caption{The self-collected multimodal multi-view dataset for gait analysis.}
  \label{fig:onecol}
\end{figure}

\begin{table}
\centering
    \setlength{\abovecaptionskip}{0.cm}
    \setlength{\belowcaptionskip}{-0.cm}
\footnotesize 
\setlength{\tabcolsep}{2.5pt} 

\begin{tabularx}{\linewidth}{c c cccc cccc}
\toprule
& & \multicolumn{2}{c}{\textbf{Front View}} & \multicolumn{2}{c}{\textbf{Left View}} & \multicolumn{2}{c}{\textbf{Right View}} & \multicolumn{2}{c}{\textbf{Back View}} \\
\cmidrule(lr){3-4} \cmidrule(lr){5-6} \cmidrule(lr){7-8} \cmidrule(lr){9-10}
\textbf{Action} & \textbf{Metric ($\downarrow$)} & \textbf{W} & \textbf{M} & \textbf{W} & \textbf{M} & \textbf{W} & \textbf{M} & \textbf{W} & \textbf{M} \\
\midrule

Walk      & \multirow{5}{*}{PA-MPJPE (mm)}
          & 97.9 & \textbf{87.6} & 105.2 & \textbf{91.6} & 108.5 & \textbf{92.3} & 102.5 & \textbf{86.3} \\
Stairs    & & 65.3  & \textbf{58.2} & 71.1  & \textbf{55.0} & 64.8  & \textbf{47.9} & 65.0  & \textbf{50.4} \\
TUG       & & 50.7  & \textbf{49.7} & \textbf{52.4}  & 53.2 & 59.0  & \textbf{53.1} & 67.5  & \textbf{62.4} \\
Bend & & 56.0  & \textbf{46.5} & 58.3  & \textbf{51.7} & 57.5  & \textbf{48.6} & \textbf{46.1}  & 47.3 \\
Squat     & & 53.2  & \textbf{47.9} & 51.9  & \textbf{46.6} & 68.6  & \textbf{49.2} & 55.3  & \textbf{46.2} \\
\midrule

\multirow{2}{*}{\begin{tabular}{@{}l@{}}Walk\end{tabular}} 

& Cadence (s)  & 0.136 & \textbf{0.102} & 0.108 & \textbf{0.101} & 0.091 & \textbf{0.088} & \textbf{0.115} & 0.119 \\
& Stride (m)   & 0.162 & \textbf{0.143} & 0.143 & \textbf{0.132} & \textbf{0.129} & 0.135 & 0.177 & \textbf{0.155} \\

\bottomrule
\end{tabularx}
\caption{Performance comparison on pose estimation error (PA-MPJPE) and gait analysis error. Lower is better for all metrics. Best result is in \textbf{bold}. W: WHAM, M: \name.}
\label{tab:final_integrated_compact}
\end{table}



\subsection{Robustness}
To assess robustness under challenging real-world conditions, we compare \name with WHAM using PA-MPJPE in three scenarios: fisheye distortion, partial occlusion, and low light.

Fisheye distortion is evaluated on EgoHumans~\cite{khirodkar2023egohumans}; partial occlusion and low-light settings are evaluated on our dataset. Table~\ref{table: Robustness} shows that \name consistently outperforms WHAM, reducing PA-MPJPE from 85 mm to 69 mm under fisheye distortion, from 50 mm to 43 mm under partial occlusion, and from 32 mm to 28 mm in low light.
Overall, results show our method maintains high pose accuracy under diverse real-world challenges. The largest gains appear under fisheye and low-light conditions, where appearance cues are unreliable and our geometric view-invariant features help suppress upstream detector errors~\cite{Xu2022ViTPoseSV}. This indicates that MoViD remains effective when appearance quality degrades and geometric consistency becomes more important.
The improvement under partial occlusion further suggests that the disentangled motion representation is less sensitive to missing or degraded local observations.



\begin{table}
    \centering
        \setlength{\abovecaptionskip}{0.cm}
    \setlength{\belowcaptionskip}{0.cm}
    \setlength{\tabcolsep}{10pt}
    \begin{tabular}{lccc}
    \toprule
    \textbf{Method} & \textbf{Fisheye} 
                    & \textbf{Partial occlusion} 
                    & \textbf{Low light} \\
    \midrule
    WHAM & 85mm & 50mm & 32mm \\
    Ours & \textbf{69mm} & \textbf{43mm} & \textbf{28mm} \\
    \bottomrule
    \end{tabular}

            \caption{Robustness across different conditions (PA-MPJPE error).}
                    \label{table: Robustness}
\end{table}

\section{Discussion}

\textbf{Analysis of success and failure cases.}
\name gains the most on datasets with diverse viewpoint coverage, such as HuMMan, where motion-view disentanglement can better leverage large cross-view variation. Improvements on 3DPW are smaller but consistent, while top-down and near-overhead views remain the most challenging because of self-occlusion and compressed body geometry.

\textbf{Lessons learned.} A throughput of 15 FPS is sufficient for our current targets, but higher frame rates would further benefit latency-critical applications. Our case studies show that adaptive flip refinement is most useful under UAV viewpoint changes, while cross-view robustness reduces the need for precise calibration in gait analysis.

\textbf{Future directions.} We will explore temporal filtering and motion priors to better handle rapid viewpoint shifts. Possible directions for improving system efficiency include lower input resolution, lighter 2D detectors, and next-generation edge hardware. We will also study additional modalities, such as neuromorphic sensors and event-based cameras.

\section{Conclusion}
\label{sec:conclusion}

We propose \name, a framework for viewpoint-invariant 3D human pose estimation that disentangles motion and view features. \name employs a view estimator to extract viewpoint information from intermediate pose features, and disentangle view-related and motion-related components through orthogonal feature projection. To further enhance robustness of motion features, \name integrates a physics-strengthened contrastive alignment mechanism that enforces motion consistency across different viewpoints, and an adaptive flip refinement module to reduce inference latency.
These results suggest that explicit motion-view disentanglement is a practical design principle for reliable pose sensing under unconstrained viewpoints.
Extensive experiments on public datasets and two self-collected  multi-view datasets demonstrate that \name achieves superior accuracy of pose estimation under diverse and challenging conditions, while maintaining real-time performance suitable for edge deployment.

\begin{acks}
This work is supported by the Research Grants Council (RGC) of Hong Kong, China, under grant ECS 26200825, the HKUST – HKUST(GZ) Cross-campus Research Collaboration “1+1+1” Joint Funding Program under G\_2025\_052, and is partly funded by the HKUST Institute for Emerging Market Studies with support from EY, under grant IEMS25EG01.
\end{acks}

\clearpage

\bibliographystyle{ACM-Reference-Format}
\bibliography{reference}

@inproceedings{cai2022humman,
  title={{HuMMan}: Multi-modal 4d human dataset for versatile sensing and modeling},
  author={Cai, Zhongang and Ren, Daxuan and Zeng, Ailing and Lin, Zhengyu and Yu, Tao and Wang, Wenjia and Fan,
          Xiangyu and Gao, Yang and Yu, Yifan and Pan, Liang and Hong, Fangzhou and Zhang, Mingyuan and
          Loy, Chen Change and Yang, Lei and Liu, Ziwei},
  booktitle={17th European Conference on Computer Vision, Tel Aviv, Israel, October 23--27, 2022,
             Proceedings, Part VII},
  pages={557--577},
  year={2022},
  organization={Springer}
}

@inproceedings{ouyang2024admarker,
  title={ADMarker: A Multi-Modal Federated Learning System for Monitoring Digital Biomarkers of Alzheimer's Disease},
  author={Ouyang, Xiaomin and Shuai, Xian and Li, Yang and Pan, Li and Zhang, Xifan and Fu, Heming and Cheng, Sitong and Wang, Xinyan and Cao, Shihua and Xin, Jiang and others},
  booktitle={Proceedings of the 30th Annual International Conference on Mobile Computing and Networking},
  pages={404--419},
  year={2024}
}

@inproceedings{do2025skateformer,
  title={Skateformer: skeletal-temporal transformer for human action recognition},
  author={Do, Jeonghyeok and Kim, Munchurl},
  booktitle={European Conference on Computer Vision},
  pages={401--420},
  year={2025},
  organization={Springer}
}

@InProceedings{shin2023wham,  
title={WHAM: Reconstructing World-grounded Humans with Accurate 3D Motion},
author={Shin, Soyong and Kim, Juyong and Halilaj, Eni and Black, Michael J.},  
booktitle={Computer Vision and Pattern Recognition (CVPR)},  
year={2024}  
}

@inproceedings{shahroudy2016ntu,
  title={NTU RGB+D: A large scale dataset for 3D human activity analysis},
  author={Shahroudy, Amir and Liu, Jun and Ng, Tian-Tsong and Wang, Gang},
  booktitle={Proceedings of the IEEE conference on computer vision and pattern recognition},
  pages={1010--1019},
  year={2016}
}

@article{liu2020ntu,
  title={NTU RGB+D 120: A large-scale benchmark for 3D human activity understanding},
  author={Liu, Jun and Shahroudy, Amir and Perez, Mauricio and Wang, Gang and Duan, Ling-Yu and Kot, Alex C},
  journal={IEEE Transactions on Pattern Analysis and Machine Intelligence},
  volume={42},
  number={10},
  pages={2684--2701},
  year={2020}
}

@INPROCEEDINGS{WEPDTOF-Pose,
  author={Huang, Linzhi and Li, Yulong and Tian, Hongbo and Yang, Yue and Li, Xiangang and Deng, Weihong and Ye, Jieping},
  booktitle={2023 IEEE/CVF Conference on Computer Vision and Pattern Recognition (CVPR)}, 
  title={Semi-Supervised 2D Human Pose Estimation Driven by Position Inconsistency Pseudo Label Correction Module}, 
  year={2023},
  pages={693-703},
  doi={10.1109/CVPR52729.2023.00074}}

@inbook{SMPL,
author = {Loper, Matthew and Mahmood, Naureen and Romero, Javier and Pons-Moll, Gerard and Black, Michael J.},
title = {SMPL: A Skinned Multi-Person Linear Model},
year = {2023},
isbn = {9798400708978},
publisher = {Association for Computing Machinery},
address = {New York, NY, USA},
edition = {1},
url = {https://doi.org/10.1145/3596711.3596800},
booktitle = {Seminal Graphics Papers: Pushing the Boundaries, Volume 2},
articleno = {88},
numpages = {16}
}

@INPROCEEDINGS{SURREAL,
  author={Varol, Gül and Romero, Javier and Martin, Xavier and Mahmood, Naureen and Black, Michael J. and Laptev, Ivan and Schmid, Cordelia},
  booktitle={2017 IEEE Conference on Computer Vision and Pattern Recognition (CVPR)}, 
  title={Learning from Synthetic Humans}, 
  year={2017},
  volume={},
  number={},
  pages={4627-4635},
  doi={10.1109/CVPR.2017.492}}

@INPROCEEDINGS{PanopTOP,
  author={Garau, Nicola and Martinelli, Giulia and Bródka, Piotr and Bisagno, Niccolò and Conci, Nicola},
  booktitle={2021 IEEE/CVF International Conference on Computer Vision Workshops (ICCVW)}, 
  title={PanopTOP: a framework for generating viewpoint-invariant human pose estimation datasets}, 
  year={2021},
  volume={},
  number={},
  pages={234-242},
  doi={10.1109/ICCVW54120.2021.00031}}

@misc{NToP,
      title={NToP: NeRF-Powered Large-scale Dataset Generation for 2D and 3D Human Pose Estimation in Top-View Fisheye Images}, 
      author={Jingrui Yu and Dipankar Nandi and Roman Seidel and Gangolf Hirtz},
      year={2024},
      eprint={2402.18196},
      archivePrefix={arXiv},
      primaryClass={cs.CV},
      url={https://arxiv.org/abs/2402.18196}, 
}

@inproceedings{mildenhall2020nerf,
  title={NeRF: Representing Scenes as Neural Radiance Fields for View Synthesis},
  author={Ben Mildenhall and Pratul P. Srinivasan and Matthew Tancik and Jonathan T. Barron and Ravi Ramamoorthi and Ren Ng},
  year={2020},
  booktitle={ECCV},
}

@article{dong2022viewfool,
  title={Viewfool: Evaluating the robustness of visual recognition to adversarial viewpoints},
  author={Dong, Yinpeng and Ruan, Shouwei and Su, Hang and Kang, Caixin and Wei, Xingxing and Zhu, Jun},
  journal={Advances in Neural Information Processing Systems},
  volume={35},
  pages={36789--36803},
  year={2022}
}

@inproceedings{ObjectNet,
 author = {Barbu, Andrei and Mayo, David and Alverio, Julian and Luo, William and Wang, Christopher and Gutfreund, Dan and Tenenbaum, Josh and Katz, Boris},
 booktitle = {Advances in Neural Information Processing Systems},
 editor = {H. Wallach and H. Larochelle and A. Beygelzimer and F. d\textquotesingle Alch\'{e}-Buc and E. Fox and R. Garnett},
 pages = {},
 publisher = {Curran Associates, Inc.},
 title = {ObjectNet: A large-scale bias-controlled dataset for pushing the limits of object recognition models},
 url = {https://proceedings.neurips.cc/paper_files/paper/2019/file/97af07a14cacba681feacf3012730892-Paper.pdf},
 volume = {32},
 year = {2019}
}

@article{Ruan2023TowardsVV,
  title={Towards Viewpoint-Invariant Visual Recognition via Adversarial Training},
  author={Shouwei Ruan and Yinpeng Dong and Han Su and Jianteng Peng and Ning Chen and Xingxing Wei},
  journal={2023 IEEE/CVF International Conference on Computer Vision (ICCV)},
  year={2023},
  pages={4686-4696},
  url={https://api.semanticscholar.org/CorpusID:259991124}
}

@inproceedings{Ruan2024OmniviewTuningBV,
  title={Omniview-Tuning: Boosting Viewpoint Invariance of Vision-Language Pre-training Models},
  author={Shouwei Ruan and Yinpeng Dong and Hanqing Liu and Yao Huang and Hang Su and Xingxing Wei},
  booktitle={European Conference on Computer Vision},
  year={2024},
  url={https://api.semanticscholar.org/CorpusID:269214060}
}

@article{Xu2022ViTPoseSV,
  title={ViTPose: Simple Vision Transformer Baselines for Human Pose Estimation},
  author={Yufei Xu and Jing Zhang and Qiming Zhang and Dacheng Tao},
  journal={ArXiv},
  year={2022},
  volume={abs/2204.12484},
  url={https://api.semanticscholar.org/CorpusID:248392410}
}

@article{Ionescu2014Human36MLS,
  title={Human3.6M: Large Scale Datasets and Predictive Methods for 3D Human Sensing in Natural Environments},
  author={Catalin Ionescu and Dragos Papava and Vlad Olaru and Cristian Sminchisescu},
  journal={IEEE Transactions on Pattern Analysis and Machine Intelligence},
  year={2014},
  volume={36},
  pages={1325-1339},
  url={https://api.semanticscholar.org/CorpusID:4244548}
}

@article{LIU2024human-robot,
title = {A skeleton-based assembly action recognition method with feature fusion for human-robot collaborative assembly},
journal = {Journal of Manufacturing Systems},
volume = {76},
pages = {553-566},
year = {2024},
issn = {0278-6125},
doi = {https://doi.org/10.1016/j.jmsy.2024.08.019},
url = {https://www.sciencedirect.com/science/article/pii/S0278612524001821},
author = {Daxin Liu and Yu Huang and Zhenyu Liu and Haoyang Mao and Pengcheng Kan and Jianrong Tan},
}

@inproceedings{goel2023humans,
    title={Humans in 4{D}: Reconstructing and Tracking Humans with Transformers},
    author={Goel, Shubham and Pavlakos, Georgios and Rajasegaran, Jathushan and Kanazawa, Angjoo and Malik, Jitendra},
    booktitle={ICCV},
    year={2023}
}

@Inproceedings{wang23refit,
  Title          = {ReFit: Recurrent Fitting Network for 3D Human Recovery},
  Author         = {Yufu Wang and Kostas Daniilidis},
  Booktitle      = {International Conference on Computer Vision},
  Year           = {2023}
}

@InProceedings{3dpw,
author="von Marcard, Timo
and Henschel, Roberto
and Black, Michael J.
and Rosenhahn, Bodo
and Pons-Moll, Gerard",
editor="Ferrari, Vittorio
and Hebert, Martial
and Sminchisescu, Cristian
and Weiss, Yair",
title="Recovering Accurate 3D Human Pose in the Wild Using IMUs and a Moving Camera",
booktitle="Computer Vision -- ECCV 2018",
year="2018",
publisher="Springer International Publishing",
address="Cham",
pages="614--631",
isbn="978-3-030-01249-6"
}

@InProceedings{rich,
    author    = {Huang, Chun-Hao P. and Yi, Hongwei and H\"oschle, Markus and Safroshkin, Matvey and Alexiadis, Tsvetelina and Polikovsky, Senya and Scharstein, Daniel and Black, Michael J.},
    title     = {Capturing and Inferring Dense Full-Body Human-Scene Contact},
    booktitle = {Proceedings of the IEEE/CVF Conference on Computer Vision and Pattern Recognition (CVPR)},
    month     = {June},
    year      = {2022},
    pages     = {13274-13285}
}

@InProceedings{kaufman2023emdb,
    author    = {Choi, Hongsuk and Moon, Gyeongsik and Chang, Ju Yong and Lee, Kyoung Mu},
    title     = {Beyond Static Features for Temporally Consistent 3D Human Pose and Shape From a Video},
    booktitle = {Proceedings of the IEEE/CVF Conference on Computer Vision and Pattern Recognition (CVPR)},
    month     = {June},
    year      = {2021},
    pages     = {1964-1973}
}

@inproceedings{AMASS:2019,
  title={AMASS: Archive of Motion Capture as Surface Shapes},
  author={Mahmood, Naureen and Ghorbani, Nima and F. Troje, Nikolaus and Pons-Moll, Gerard and Black, Michael J.},
  booktitle = {The IEEE International Conference on Computer Vision (ICCV)},
  year={2019},
  month = {Oct},
  url = {https://amass.is.tue.mpg.de},
  month_numeric = {10}
}

@InProceedings{InstaVariety,
  title={Learning 3D Human Dynamics from Video},
  author = {Angjoo Kanazawa and Jason Y. Zhang and Panna Felsen and Jitendra Malik},
  booktitle={Computer Vision and Pattern Recognition (CVPR)},
  year={2019}}

@inproceedings{mono-3dhp2017,
 author = {Mehta, Dushyant and Rhodin, Helge and Casas, Dan and Fua, Pascal and Sotnychenko, Oleksandr and Xu, Weipeng and Theobalt, Christian},
 title = {Monocular 3D Human Pose Estimation In The Wild Using Improved CNN Supervision},
 booktitle = {3D Vision (3DV), 2017 Fifth International Conference on},
 url = {http://gvv.mpi-inf.mpg.de/3dhp_dataset},
 year = {2017},
 organization={IEEE},
 doi={10.1109/3dv.2017.00064},
}

@misc{xia2025reconstructinghumansbiomechanicallyaccurate,
      title={Reconstructing Humans with a Biomechanically Accurate Skeleton}, 
      author={Yan Xia and Xiaowei Zhou and Etienne Vouga and Qixing Huang and Georgios Pavlakos},
      year={2025},
      eprint={2503.21751},
      archivePrefix={arXiv},
      primaryClass={cs.CV},
      url={https://arxiv.org/abs/2503.21751}, 
}

@inproceedings{Song2020Pose2Pose,
  title={Pose2Pose: Human Pose Transfer by Attention-Based Keypoint Graph},
  author={Song, Yuming and Zhang, Wei and Chang, Xiaojun and Zhang, Yi and Liu, Wei and Wu, Yao and Wang, Lei},
  booktitle={European Conference on Computer Vision (ECCV)},
  year={2020},
  pages={236--252}
}

@inproceedings{Guler2018DensePose,
  title={DensePose: Dense Human Pose Estimation In-the-Wild},
  author={Guler, Riza Alp and Neverova, Natalia and Kokkinos, Iasonas},
  booktitle={Proceedings of the IEEE/CVF Conference on Computer Vision and Pattern Recognition (CVPR)},
  year={2018},
  pages={7297--7306}
}

@inproceedings{Kolotouros2019SPIN,
  title={Learning to Reconstruct 3D Human Pose and Shape via Model-Fitting in the Loop},
  author={Kolotouros, Nikos and Pavlakos, Georgios and Black, Michael J and Daniilidis, Kostas},
  booktitle={Proceedings of the IEEE International Conference on Computer Vision (ICCV)},
  year={2019},
  pages={2252--2261}
}

@inproceedings{Kolotouros2019GraphCMR,
  title={Convolutional Mesh Regression for Single-Image Human Shape Reconstruction},
  author={Kolotouros, Nikos and Pavlakos, Georgios and Daniilidis, Kostas},
  booktitle={Proceedings of the IEEE Conference on Computer Vision and Pattern Recognition (CVPR)},
  year={2019},
  pages={4501--4510}
}

@article{stenum2021applications,
  title={Applications of Pose Estimation in Human Health and Performance across the Lifespan},
  author={Stenum, J. and Cherry-Allen, K. M. and Pyles, C. O. and Reetzke, R. D. and Vignos, M. F. and Roemmich, R. T.},
  journal={Sensors},
  volume={21},
  number={21},
  pages={7315},
  year={2021},
  publisher={MDPI},
  doi={10.3390/s21217315}
}

@article{hannan2021portable,
  author = {Hannan, Abdul and others},
  title = {A Portable Smart Fitness Suite for Real-Time Exercise Monitoring and Posture Correction},
  journal = {Sensors},
  volume = {21},
  number = {19},
  pages = {6692},
  year = {2021},
  month = {10},
  day = {8},
  doi = {10.3390/s21196692}
}

@INPROCEEDINGS{falldetection,
  author={Mahesh, Ghattamaneni and Kalidas, M.},
  booktitle={2023 International Conference on Advances in Computation, Communication and Information Technology (ICAICCIT)}, 
  title={A Real-Time IoT Based Fall Detection and Alert System for Elderly}, 
  year={2023},
  volume={},
  number={},
  pages={327-331},
  doi={10.1109/ICAICCIT60255.2023.10465914}}

@ARTICLE{yolofalldetection,
  author={Priadana, Adri and Nguyen, Duy-Linh and Vo, Xuan-Thuy and Choi, Jehwan and Ashraf, Russo and Jo, Kanghyun},
  journal={IEEE Access}, 
  title={HFD-YOLO: Improved YOLO Network Using Efficient Attention Modules for Real-Time One-Stage Human Fall Detection}, 
  year={2025},
  volume={13},
  number={},
  pages={41248-41258},
  doi={10.1109/ACCESS.2025.3547360}}

@article{Men2023FocalizedCV,
  title={Focalized Contrastive View-invariant Learning for Self-supervised Skeleton-based Action Recognition},
  author={Qianhui Men and Edmond S. L. Ho and Hubert P. H. Shum and Howard Leung},
  journal={Neurocomputing},
  year={2023},
  volume={537},
  pages={198-209},
  url={https://api.semanticscholar.org/CorpusID:257899093}
}

@ARTICLE{PVNet,
  author={Peng, Sida and Zhou, Xiaowei and Liu, Yuan and Lin, Haotong and Huang, Qixing and Bao, Hujun},
  journal={IEEE Transactions on Pattern Analysis and Machine Intelligence}, 
  title={PVNet: Pixel-Wise Voting Network for 6DoF Object Pose Estimation}, 
  year={2022},
  volume={44},
  number={6},
  pages={3212-3223},
  doi={10.1109/TPAMI.2020.3047388}}

@Article{s19091988,
AUTHOR = {Martínez-Villaseñor, Lourdes and Ponce, Hiram and Brieva, Jorge and Moya-Albor, Ernesto and Núñez-Martínez, José and Peñafort-Asturiano, Carlos},
TITLE = {UP-Fall Detection Dataset: A Multimodal Approach},
JOURNAL = {Sensors},
VOLUME = {19},
YEAR = {2019},
NUMBER = {9},
ARTICLE-NUMBER = {1988},
URL = {https://www.mdpi.com/1424-8220/19/9/1988},
PubMedID = {31035377},
ISSN = {1424-8220},
DOI = {10.3390/s19091988}
}

@incollection{hazewinkel2001orthogonalization,
  author       = {Hazewinkel, Michiel},
  title        = {Orthogonalization},
  booktitle    = {Encyclopaedia of Mathematics},
  publisher    = {Springer},
  year         = {2001},
  isbn         = {978-1-55608-010-4},
  editor       = {Hazewinkel, Michiel}
}

@inproceedings{sun2018integral,
  title={Integral human pose regression},
  author={Sun, Xiao and Xiao, Bin and Wei, Fangyin and Liang, Shuang and Wei, Yichen},
  booktitle={Proceedings of the European Conference on Computer Vision (ECCV)},
  pages={529--545},
  year={2018}
}

@article{khirodkar2023egohumans,
  title={EgoHumans: An Egocentric 3D Multi-Human Benchmark},
  author={Khirodkar, Rawal and Bansal, Aayush and Ma, Lingni and Newcombe, Richard and Vo, Minh and Kitani, Kris},
  journal={arXiv preprint arXiv:2305.16487},
  year={2023}
}

@Article{cite_skeleton_orientation,
AUTHOR = {Sun, Haixun and Zhang, Yanyan and Zheng, Yijie and Luo, Jianxin and Pan, Zhisong},
TITLE = {G2O-Pose: Real-Time Monocular 3D Human Pose Estimation Based on General Graph Optimization},
JOURNAL = {Sensors},
VOLUME = {22},
YEAR = {2022},
NUMBER = {21},
ARTICLE-NUMBER = {8335},
URL = {https://www.mdpi.com/1424-8220/22/21/8335},
PubMedID = {36366035},
ISSN = {1424-8220},
DOI = {10.3390/s22218335}
}

@inproceedings{Nie2019SPM,
  author    = {Xuecheng Nie and Jiashi Feng and Jianfeng Zhang and Shuicheng Yan},
  title     = {Single-Stage Multi-Person Pose Machines},
  booktitle = {Proceedings of the IEEE/CVF International Conference on Computer Vision (ICCV)},
  year      = {2019},
  pages     = {6950--6959},
  doi       = {10.1109/ICCV.2019.00705}
}

@inproceedings{Li2022SimCC,
  author    = {Yanjie Li and Sen Yang and Peidong Liu and Shoukui Zhang and Yunxiao Wang and Zhicheng Wang and Wankou Yang and Shu-Tao Xia},
  title     = {{SimCC}: A Simple Coordinate Classification Perspective for Human Pose Estimation},
  booktitle = {Proceedings of the European Conference on Computer Vision (ECCV)},
  year      = {2022},
  pages     = {89--106},
  publisher = {Springer},
  series    = {Lecture Notes in Computer Science},
  volume    = {13666},
  doi       = {10.1007/978-3-031-20068-7_6}
}

@inproceedings{mhformer,
  title={MHFormer: Multi-Hypothesis Transformer for 3D Human Pose Estimation},
  author={Li, Wenhao and Liu, Hong and Tang, Hao and Wang, Pichao and Van Gool, Luc},
  booktitle={Proceedings of the IEEE/CVF Conference on Computer Vision and Pattern Recognition (CVPR)},
  pages={13147--13156},
  year={2022}
}

@inproceedings{repnet,
  title={RepNet: Weakly Supervised Training of an Adversarial Reprojection Network for 3D Human Pose Estimation},
  author={Wandt, Bastian and Rosenhahn, Bodo},
  booktitle={Computer Vision and Pattern Recognition (CVPR)},
  month={June},
  year={2019}
}

@inproceedings{Ghezelghieh2016CNN3DV,
  title={Learning Camera Viewpoint Using CNN to Improve 3D Body Pose Estimation},
  author={Ghezelghieh, Maryam Fathollahi and Kasturi, Rangachar and Sarkar, Sudeep},
  booktitle={Proc. IEEE International Conference on 3D Vision (3DV)},
  year={2016}
}

@inproceedings{Wang2020PredictingCV,
  title={Predicting Camera Viewpoint Improves Cross-Dataset Generalization for 3D Human Pose Estimation},
  author={Wang, Zhe and Shin, Daeyun and Fowlkes, Charless C.},
  booktitle={Proc. European Conference on Computer Vision (ECCV) Workshops},
  year={2020}
}

@inproceedings{Sun2019HRNet,
  title={Deep High-Resolution Representation Learning for Human Pose Estimation},
  author={Sun, Ke and Xiao, Bin and Liu, Dong and Wang, Jingdong},
  booktitle={CVPR},
  year={2019}
}

\clearpage

\end{document}